\documentclass{article}

\usepackage[preprint]{neurips_2024}

\usepackage[utf8]{inputenc} % allow utf-8 input
\usepackage[T1]{fontenc}    % use 8-bit T1 fonts
\usepackage{hyperref}       % hyperlinks
\usepackage{url}            % simple URL typesetting
\usepackage{booktabs}       % professional-quality tables
\usepackage{amsfonts}       % blackboard math symbols
\usepackage{nicefrac}       % compact symbols for 1/2, etc.
\usepackage{microtype}      % microtypography
\usepackage{xcolor}         % colors
\usepackage{amsmath}
\usepackage{graphicx}
\usepackage{algorithm}
\usepackage{algorithmic}
\usepackage{multirow} 
\usepackage{amsmath}
\usepackage{amssymb}
\usepackage{wrapfig,lipsum,booktabs}
\usepackage[bottom]{footmisc}

\title{Enhancing GNNs Performance on Combinatorial Optimization by Recurrent Feature Update}% of Vertex States}

\author{ %
  Daria Pugacheva \\
   JIHT RAS, \\ 
   \texttt{pugachyova.d@gmail.com}\\ 
  \And
  Andrei Ermakov \\
  HSE University \\
  \And
  Igor Lyskov \\
  RMIT \\
  \And
  Ilya Makarov \\
  AIRI, MEPhI \\ 
  \And
  Yuriy Zotov \\
  Independent Researcher \\ 
}

\begin{document}

\maketitle

\begin{abstract}
Combinatorial optimization (CO) problems are crucial in various scientific and industrial applications. Recently, researchers have proposed using unsupervised Graph Neural Networks (GNNs) to address NP-hard combinatorial optimization problems, which can be reformulated as Quadratic Unconstrained Binary Optimization (QUBO) problems. GNNs have demonstrated high performance with nearly linear scalability and significantly outperformed classic heuristic-based algorithms in terms of computational efficiency 
% and resource utilization 
on large-scale problems. 
% In simple words, each node in a QUBO graph is classified into two classes based on whether the node is a part of solution. 
However, when utilizing standard node features, GNNs 
% suffer from overfitting and 
tend to get trapped to suboptimal local minima of the energy landscape, resulting in low quality solutions.
We introduce a novel algorithm,  denoted hereafter as QRF-GNN, leveraging the power of GNNs to efficiently solve CO problems with QUBO formulation. It relies on unsupervised learning by minimizing the loss function derived from QUBO relaxation. The proposed key components of the architecture include the recurrent use of intermediate GNN predictions, parallel convolutional layers and combination of static node features as input. Altogether, it helps to adapt the intermediate solution candidate to minimize QUBO-based loss function, taking into account not only static graph features, but also intermediate predictions treated as dynamic, i.e. iteratively changing recurrent features. 
The performance of the proposed algorithm has been evaluated on the canonical benchmark datasets for maximum cut, graph coloring and maximum independent set problems. Results of experiments show that QRF-GNN drastically surpasses existing learning-based approaches and is comparable to the state-of-the-art conventional heuristics, improving their scalability on large instances.
\end{abstract}

\section{Introduction}
\label{sec:introduction}
% by p_xiaoluban
%Graph neural networks (GNNs) have emerged as a powerful tool for solving combinatorial optimization problems.
%These problems arise in a wide range of applications, such as logistics, scheduling, and resource allocation.
%Traditionally, solving combinatorial optimization problems requires designing specialized algorithms that can handle
%the specific problem at hand.
%However, GNNs offer a more general approach that can be applied to a wide range of problems.
%In this article, we will explore the use of GNNs for solving Max-Cut, MIS and graph coloring problems and
%advantages of GNNs over traditional methods, and some recent advances in the field.
%We will also discuss some of the challenges that remain in using GNNs for combinatorial optimization and potential
%directions for future research.

Combinatorial Optimization (CO) is a well-known subject in computer science, bridging operations research, discrete mathematics and optimization.
Informally, given some ground set, the CO problem is to select the combination of its elements, such that it lies on the problem's feasible domain and the cost
of this combination is minimized.
A significant amount of CO problems are known to be NP-hard, meaning that they are computationally intractable under ``$P\not=NP$'' conjecture
and the scope of application for exact algorithms to solve them is very narrow. Therefore, the development of heuristic methods that provide high-accuracy solutions in acceptable amount of time is a crucial challenge in the field ~\cite{BOUSSAID201382}.

A wide range of CO problems are defined on structural data, and their solutions are encoded as a subset of graph's edges or nodes. 
%Many application domains can be formulated on graphs such as logistics, resource allocation, scheduling, planning, etc. 
Implicit regularities and patterns often arise in graph structure and features, making the use of machine learning and especially graph neural networks (GNNs) very promising~\cite{Cappart23}. %GNNs are able to leverage the topological and feature-based information inherent in graphs, providing a powerful tool for learning and optimization in CO problems. 
\\ \\
One notable CO problem is the Quadratic Unconstrained Binary Optimization (QUBO), which aims to minimize a pseudo-Boolean polynomial $\mathcal{F}(x)$ of degree two~\cite{Hammer1969, Boros1991TheMP}:
\begin{equation}
    \label{eq:qubo}
    \begin{split}
        & \min_{x \in \{0,1\}^n} \mathcal{F}(x) = \sum_{i=1}^n{\sum_{j=1}^n x_iA_{ij}x_j} + \sum_{i=1}^n c_ix_i = x^TQx,
    \end{split}
\end{equation}
where the symmetric matrix of coefficients $A \in \mathbb{R}^{n \times n}$ and $c \in \mathbb{R}^n$ encode the initial problem,  $x =  (x_1,x_2,\dots,x_n)^T$ is the vector of binary variables $x_i \in \{0,1\}$ and $Q \in \mathbb{R}^{n \times n}  = A + \text{diag(c)} $. The latter equivalence in Equation \ref{eq:qubo} holds because $x_{i}^2 = x_{i}$ for all $i$. 

Despite the QUBO problem has been studied for a very long time \cite{Hammer1970SomeRO}, it has recently attracted much attention as a way to formulate other CO problems ~\cite{Glover2022, Lucas2014}, mainly due to the emerging interest in the development of quantum computational devices \cite{Boixo2013,CIM}. By applying simple reformulation techniques, such as constraint penalization \cite{alice1997}, a huge number of CO problems can be formulated as QUBO, which makes algorithms for its solution especially valuable in practice.

% evolve into its ground state and the solution to the encoded problem is achieved.

Researchers from Amazon ~\cite{Schuetz2022} proposed applying GNN to solve QUBO in unsupervised manner, using the differentiable continuous relaxation of Equation \ref{eq:qubo} as a loss function. 
This method does not require training data, meaning that GNN is trained to solve the particular problem instance end-to-end, and that it can be considered as autonomous heuristic algorithm. The QUBO matrix is associated with its adjacency graph, on which the GNN is trained. Therefore, such an approach allows using a graph network to solve a wide variety of problems formulated as QUBO, and is not limited to combinatorial problems initially defined on graphs. The main advantage of unsupervised GNNs is the ability to solve very large-scale problems. However, their ability to obtain accurate solutions has been questioned in the community~\cite{Boettcher2023,Angelini2023}: it was claimed that even greedy algorithms outperform GNN. ~\cite{wang2023unsupervised} suggested that it happens because GNNs get stuck in local optima when trained for particular problem instances. 
To overcome these limitations, we introduce a novel unsupervised QUBO-based Graph Neural Network with a Recurrent Feature (QRF-GNN). 
%The existing  main idea is to overcome limitations of existing architectures.... and bring the power of ... to achieve superior performance while retaining the quality of ... .
%We provide results of numerical experiments on popular benchmark instances of maximum cut (Max-Cut), graph coloring and maximum independent set (MIS) problems.
%We show that the recurrent feature drastically improves the performance of GNNs owing to information about the current status of each node in the graph.
%A combination of artificial input features and parallel graph convolutional layers is offered for additional benefit.
%We show that the proposed method outperforms existing state-of-the-art (SOTA) leaarning based approaches for the considered problems.
The main contribution of our work are as follows:
\begin{itemize}
\item We propose a novel GNN architecture QRF-GNN based on a new type of recurrent node connection, which is able to provide accurate solutions for large-scale CO. 
\item
We show that the proposed recurrent design significantly improves the performance of GNNs for all considered types of convolutions. 
\item

%We conduct experiments on maximum cut (Max-Cut), graph coloring and maximum independent set (MIS) problems. 

We show that our QRF-GNN outperforms existing SOTA learning based approaches for all benchmarks of maximum cut (Max-Cut), graph coloring and maximum independent set (MIS) problems. It also competes with the best existing conventional heuristics, while improving them in scalability and computational time.

%The results underpin the potential of QRF-GNN as an effective technique for CO, offering competitive performance with the best existing conventional heuristics.
\end{itemize}

%taking less time to find solutions, especially for large graphs.

%We provide results of numerical experiments on popular benchmark instances of maximum cut (Max-Cut), graph coloring and maximum independent set (MIS) problems. We show that the proposed method outperforms existing state-of-the-art (SOTA) leaarning based approaches for the considered problems. The results underpin the potential of QRF-GNN as an effective techinique for CO, offering competitive performance with the best existing conventional heuristics.

%\section{Background}
%\label{sec:background}
%\input{background.tex}

\section{Proposed QRF-GNN method and Experiment Design}
\label{sec:rec-pi-gnn}
% In this section, we provide a background on QUBO solution using GNNs, formulate our novel method for recurrent feature aggregation which is the main contribution of this paper, describe in details its architecture and, design the experiments on three well-known CO problems on graphs
 
In this section, we first provide a background on graph neural networks and how they are applied to solve QUBO. Then, we describe a new proposed type of recurrent connection for GNNs to enhance their performance on CO. This will be followed by a detailed description of the developed QRF-GNN architecture and a selected set of static features. The last part is devoted to a description of the loss function design
% the experiments design 
for three well-known CO problems on graphs.
 
 \begin{figure*}[tb]
 	\includegraphics[width=\linewidth]{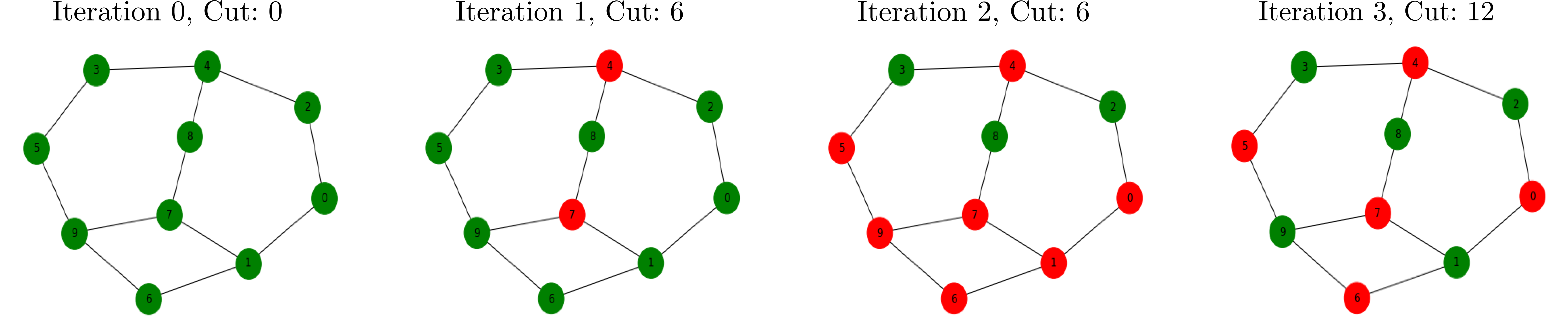}
 	\caption{The example of QRF-GNN work on a toy graph of 10 vertices and 12 edges on the Max-Cut problem. The Max-Cut involves partitioning of graph's nodes into two sets such that the number of edges between sets is maximized. At each iteration, the rounded (discretized) solution is shown: red color refers to $x_i = 0$, green to $x_i = 1$. Using the recurrent feature, QRF-GNN aims to reclassify each node to the opposite class of the majority of its nearest neighbors at each iteration, thereby optimizing the cut. On this instance, QRF-GNN is able to achieve the optimal solution in just a few iterations, while non-recurrent architectures (e.g., PI-GNN) require $\sim50$ times more iterations.
  }
 	\label{fig:6iters}
 \end{figure*}

\subsection{Graph Neural Networks for Combinatorial Optimization}\label{subsec:rec1}
%The method to which this paper is devoted is based on the incorporation of recurrence into graph neural networks.
Graph neural networks are capable to learn complex graph-structured data by capturing relational information.
During training process, each of the nodes is associated with a vector which is updated based on the information from neighboring nodes.

Let us consider an undirected graph $G=(V, E)$ with a vertex set $V = \{1, \dots, n \}$ and an edge set $E = \{(i, j): i,j \in V\}$.
Let $h^l_i \in \mathbb{R}^{m_l}$ be a feature vector for a node $i$ and $h^l_j \in \mathbb{R}^{m_l}$ a vector for a node $j$ at the $l$-th convolution layer and let $e=(i, j)$ be an edge between nodes $i$ and $j$.
% $h^l_0 \in \mathbb{R}^{m_l}$ is typically derived from selected graph input features. 

% add intuition for choosing features input and cite some work

%We do not consider here the presence of a feature vector for edges. 

In this paper, we consider GNNs that are based on the message passing protocol
to exchange information between nodes. This protocol consists of the two main parts: message accumulation and message aggregation.
Message accumulation computes a message $m^{l+1}_e$ for an edge $e$ using a function $\phi$, which determines how information will be collected. Message passing includes aggregation of collected messages from a node $i$ neighbors by function $\rho$ and then an update of a feature vector $h^l_i$ for the node $i$ by applying function $f$ with trainable weights $W^l$. So the whole process is described as follows:
\begin{equation}
    \label{eq:message-acc}
    m^{l+1}_e = \phi \left( h^l_i, h^l_j \right), \quad h^{l + 1}_i = f \left( h^l_i, \rho \left( \{ m^{l+1}_e: (i, j) \in E \} \right) \right).
\end{equation}

There are a variety of approaches how to define input features $h^0_i$ for the node $i$. It could consist of a one-hot encoding vector of a node label, a random or shared dummy vector \cite{PosStructFeatures}, a pagerank \cite{BRIN1998107} or a degree of a node. It is also possible to use a trainable embedding layer before graph convolutions \cite{Schuetz2022}.

% :
% \begin{equation}
%     \label{eq:message_update}
%     h^{l + 1}_i = f \left( h^l_i, \rho \left( \{ m^{l+1}_e: (i, j) \in E \} \right) \right).
% \end{equation}
%Several attempts to combine recurrence with GNNs have already been made.
%The most prominent ones that can be applied to solve CO problems are GraphSAGE \cite{hamilton2017sage} and RUN-CSP \cite{Tonshoff2021}, both based on the use of LSTM \cite{LSTM}.
 To solve a particular CO problem, \cite{Schuetz2022} proposed to use the continuous relaxation of the QUBO formulation (Equation \ref{eq:loss-pi-gnn}) as a loss function for GNNs, introducing their physics-inspired GNN (PI-GNN). Replacing the binary decision variables $x_i \in \{0, 1\}$ in Equation \ref{eq:qubo} with continuous probability parameters $p_i(\theta)$ yields:
\begin{equation}
    \label{eq:loss-pi-gnn}
    \begin{split}
        &  \mathcal{F}(x) = x^T Q x \rightarrow
        \mathcal{L}(\theta) = p(\theta)^T Qp(\theta), \ \ \ p(\theta) \in \left[0, 1\right]^{n},
    \end{split}
\end{equation}
\noindent where $\theta$ are neural network parameters. After the final $N_l$-th convolution layer a softmax or sigmoid activation function is applied to compress the final embeddings $h^{N_l}$ into probabilities $p(\theta)$. 

As a result of training, GNN obtains the continuous solution $p_i(\theta)$ for each node. In order to obtain a solution of the original discrete problem, $p_i(\theta)$ has to be converted into the discrete variable $x_i$. The simplest approach is to apply an indicator function $\mathbb{I}_{p_i>p^*}$ with a threshold $p^*$, as we use in our setup.   
Alternatively, sampling discrete variables from Bernouli distribution, or greedy methods can be employed \cite{wang2022unsupervised}. 
% or assign a value of 1 to the variable with the highest probability and 0 to the others
% rounding each $p_i(\theta)$ to the nearest boolean variable:  $ x_i = \text{int}(p_i(\theta))$. 

% However, we observe that QRF-GNN almost always converges to feasible discrete solutions, therefore we apply simple rounding to obtain $x$. 

%The QUBO matrix is associated with its adjacency graph, on which the GNN is trained. Therefore, such an approach allows using a graph network to solve a wide variety of problems formulated as QUBO, and is not limited to combinatorial problems initially defined on graphs. 
% explain how to derive solution from the final x^\hat

%The base  architecture of PI-GNN consists of a trainable embedding layer to produce input features of nodes and
%several graph convolutional layers (GCN by \cite{kipf2017semisupervised} or GraphSAGE by \cite{hamilton2017sage}).

%In \cite{schuetz2022coloring} this model is applied for solving the graph coloring problem.

%In this paper, we propose a QRF-GNN framework that combines a new type of recurrent connection with graph neural network and a QUBO-based loss function \ref{eq:loss-pi-gnn}.
% performs the classification of the graph vertices that will minimize the loss function.

\subsection{Proposed QRF-GNN Recurrent Framework}\label{subsec:rec2}

\textbf{Motivation.} Existing GNNs often rely solely on the static node features, which can lead to significant limitations. 
% For example, in PI-GNN, the node input feature vector $h^0_i$ are one-hot encoded and then pass through a trainable embedding layer. 
%This standard approach causes overfitting on these features, and therefore, unsupervised GNN often result being trapped in local extrema during training 
When applying GNNs to solve a specific problem instance, this standard approach can cause the model to become biased towards these features and face node ambiguity issues, resulting in unsupervised GNNs frequently getting trapped in local extrema during training \cite{wang2023unsupervised}. Another intuition is related to the QUBO objective \ref{eq:loss-pi-gnn}. It represents the sum of the interaction coefficients of all pairs of non-zero variables. Therefore, the local optimal decision on whether $x_i = 0$ or $x_i = 1$ strongly depends on the classes of the neighbors of node $i$. Based on these two considerations, we propose to recursively use the predicted probability data from the previous iteration as an additional dynamic feature of the node, and utilize it directly through the message-passing protocol. Consequently, QRF-GNN operates as an iterative optimization process where nodes update its probabilities based on the probabilities of their nearest neighbors from the previous iteration, aiming to minimize \ref{eq:loss-pi-gnn}. This process is illustrated in Figure \ref{fig:6iters}, which shows the step-by-step work of QRF-GNN on a toy instance for the maximum cut optimization problem. Details of the computational experiment on this graph can be found in Appendix \ref{subsec:toy}.

%(Figure \ref{fig:prgnn}). 

%In order to use graph information, node features are usually generated in a way taking into account graph descriptive statistics. The problem in such an approach suggested in \cite{} an criticized in \cite{} lies in overfitting graph structure and strong initial bias in input features impacting the final solution.

%In contrast to \cite{}, we propose to recursively use the predicted probabilities data from the previous step as an additional feature of the node and utilize it directly through the message-passing protocol (Figure \ref{fig:prgnn}).

\textbf{Recurrent GNN background.} In the regular training process of GNNs, static feature vectors is used as an input to the network, then these vectors are updated through graph convolution layers, after which the loss function is computed and backpropagation with updates of GNN weights occurs. In the previous works, such as Gated Graph Convolution \cite{li2015gated} or RUN-CSP \cite{Tonshoff2021}, it has been proposed to incorporate recurrence into the GNN training. For this purpose, the LSTM or GRU cell was used to update node hidden and state  after message accumulation step, then these vectors are passed back to the input of the convolution layer. The result was a sequence of node states, the length of which is the given hyperparameter of the model. The evolution of states was performed with a fixed matrix of layer weights $W^l$, taking into account the memory of all previous states. The model weights are updated after the whole sequence is processed, then input vectors are reinitialized with static default values and the next iteration starts.

With these previous approaches, the recurrent update is performed without information about how it affects the loss function. 
There are also limitations due to the memory propagation of non-optimal previous states which can interfere with the correct abrupt change of the current state, even if this decision is optimal according to the loss function. 

\textbf{Proposed recurrent design.} In the QRF-GNN approach, at each iteration $t+1$ the output state vector $h^{t, N_l}_i$ of the node $i$ from previous iteration $t$ is recurrently used as a dynamic node feature. 

\begin{wrapfigure}{R}{0.525\textwidth}
\begin{minipage}{0.52\textwidth}
\begin{algorithm}[H]
    \caption{The QRF-GNN recurrence design}
    \label{alg:recurrence_prgnn}
    \begin{algorithmic}[1]
    \STATE{\bfseries Input:} Graph $G(V, E)$, features  $\{ a_i, \forall i \in V \}$
    \STATE{\bfseries Output:} Class probabilities $\{ p_i, \forall i \in V \}$ \\
    \FOR{$t \in \{0, \dots, N_t-1 \}$}
        \FOR{$i \in V$}
        \STATE$h^{t+1, 0}_i \gets \big[a_i, h^{t, N_l}_i \big]$
        \ENDFOR
        
        \FOR{$l \in \{0, \dots, N_l-1 \}$}
            \FOR{$i \in V$}
            \STATE $h^{t+1, l+1}_{N(i)} \gets \rho  \left( \left\{ h^{t+1, l}_j, \forall j \in N(i) \right\} \right)$
            \STATE $h^{t+1, l+1}_i \gets f \left( W^{t, l} \big[h^{t+1, l}_i \ h^{t+1, l+1}_{N(i)}\big] \right)$
            \ENDFOR
        \ENDFOR
        
        \STATE  $\mathcal{L} \gets \mathcal{L}\left[ \sigma \left( \left\{ h^{t+1, N_l}_i, \forall i \in V \right\} \right) \right]$
        
        \STATE $W^{t+1, l} \gets W^{t, l} - \gamma \nabla_{\theta^l} \mathcal{L}$, $\forall l$
    \ENDFOR
    \STATE$\{ p_i \gets \sigma(h_i^{N_t,N_l}), \forall i \in V \}$
\end{algorithmic}
\end{algorithm}
\end{minipage}
\end{wrapfigure}
Thus, the combination of static $a_i$ and dynamic recurrent features is treated as input of the graph neural network (see Algorithm \ref{alg:recurrence_prgnn}). Model weights are updated according to the loss function at each iteration $t$, so the matrix $W^l$ at each GNN layer $l$ becomes time dependent $W^{t, l}$ while processing recurrent update. In this design, the sequence of states is not accumulated with a fixed length per iteration as in the RUN-CSP or Gated Graph Convolution, and it is limited only by the number of iterations and convergence criteria. Optimization is performed for the current distribution of node states, and not for the sequence of states changes. 

Thus, the neural network can adjust states of nodes with recurrent update to minimize the loss function at each iteration $t+1$, taking into account the states of neighbors obtained at iteration $t$ and the initial static features of nodes. The proposed method allows to change the current state of a node closer to the optimal one without long-term memory inertia and therefore explore the state space more widely.

%until convergence is achieved 

%process rather than the intermediate state vectors, and the optimization is performed at each iteration for th
%The use of recurrence makes a fundamental improvement in results over the incremental contribution from the architecture (see the ablation study in the Appendix

\subsection{QRF-GNN Architecture and Training Pipeline}\label{subsec:arch}
The proposed recurrent framework admits the use of different graph convolution layers. In this work, three options are explored, namely GATv2 \cite{brody2021attentive}, GCN \cite{kipf2017semisupervised} and SAGE \cite{hamilton2017sage} convolutional layers. 
We show that the recurrent connection improves results in all cases, but the SAGE convolution attains better performance and was therefore chosen for the experiments (please see the ablation study \ref{sec:ablation_main} for more details).

To improve the representative power of GNNs, we suggest the use of parallel layers, which represents multi-level feature extraction similar to Inception module from computer vision field by \cite{inception}.
Final architecture of QRF-GNN consists of three SAGE convolutions with different types of aggregation (see Figure \ref{fig:prgnn} and detailed description in Appendix \ref{sec:setup}).

Mean and pooling aggregation functions were chosen for two parallel intermediate SAGE layers, and the mean aggregation function was chosen for the last SAGE layer.
The pool aggregation allows to determine the occurrence of certain classes among features of neighboring nodes.
The mean aggregation shows the ratio of the number of different classes in the neighborhood of the considered node.
This architecture configuration with a small number of successive layers allows to store information about local neighborhoods without much over-smoothing \cite{Rusch2023ASO}, and its advantages are supported by the ablation study in Appendix \ref{sec:ablation}.

\begin{figure}[h]
\centering
	\includegraphics[width=0.8\linewidth]{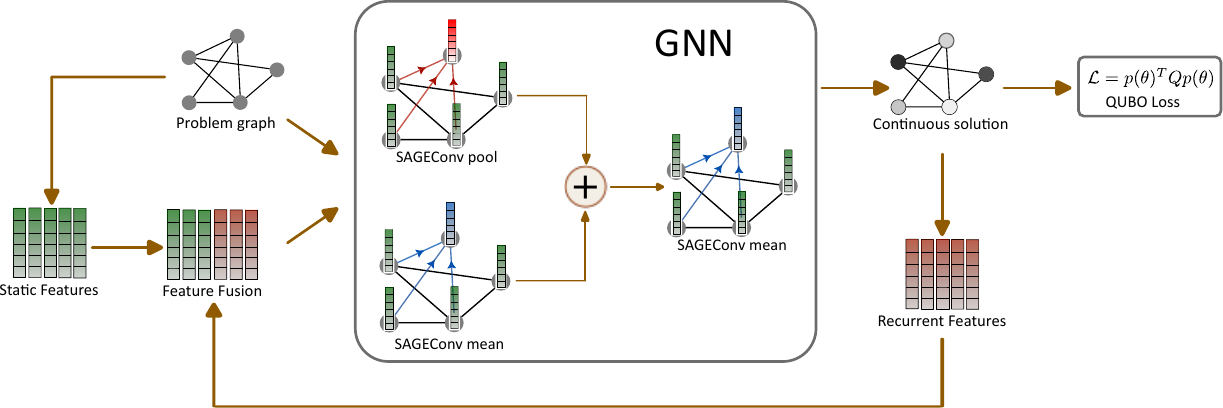}
	\caption{The QRF-GNN architecture. Firstly, the problem graph is associated with the initial QUBO problem, and the node static features are extracted. Then the dynamic recurrent features from the previous iteration is concatenated with the static features. Finally, these fused input vectors along with the graph data pass through the graph neural network to update probabilities $p_i$ in Eq. \ref{eq:loss-pi-gnn}.}
	\label{fig:prgnn}
\end{figure}

%Being an unsupervised learning method, QRF-GNN optimizes the loss function which is essentially a relaxation of the QUBO problem.
%Replacing the binary decision variables $x_i \in \{0, 1\}$ with continuous probability parameters $p_i(\vtheta)$ yields:
%\begin{equation}
%    \label{eq:loss-pi-gnn}
%    \begin{split}
%        \vx &\rightarrow \vp(\vtheta) \in \left[0, 1\right]^{{\times} \vert \gV \vert }, \\
%        F_\text{QUBO} = \vx^T \mQ \vx &\rightarrow
%        \mathcal{L}_\text{QUBO}(\vtheta) = \vp(\vtheta)^T \mQ \vp(\vtheta),
%    \end{split}
%\end{equation}
%\noindent where $\vtheta$ are neural network parameters.

%Now the trick is to recursively use the predicted probabilities data from the previous step as an additional feature of the node (see Figure~\ref{fig:prgnn})
%and directly utilize it through the message-passing protocol.
%It turns GNN training into an iterative optimization process in which the neural network adjusts the assignment of nodes classes under the influence
%of not only initial nodes properties, but also of the most probable classes of its neighbors.
%The use of recurrence makes a fundamental improvement in results over the incremental contribution from the architecture (see the ablation study in the Appendix).

%\IncMargin{1em}

% The whole forward propagation process at the time step $t$ can be described in Algorithm~\ref{alg:forward_prgnn}. The batch normalization (BN) has trainable parameters $\gamma$ and $\beta$ and is done over all nodes for each feature dimension.

As mentioned above, there are several methods how to generate static input features $a_i$, which then go through a neural network.
In this work, we create a static feature vector as a composite vector of a random part, shared vector \cite{PosStructFeatures} and pagerank \cite{BRIN1998107}.
At the first iteration, the probability vector $h^{0, 0}_i$ is initialized with zeros.
Another way involves one-hot encoding for each node and then training a special embedding layer as in the PI-GNN \cite{Schuetz2022} architecture.
It allows the neural network itself to learn the most representative features.
However, we do not use an embedding layer, since it requires additional computational resources and has shown no benefit over artificial features within the framework of conducted experiments (see Appendix \ref{sec:ablation}).

% It should be noted that the use of recurrence makes a fundamental improvement in results over the incremental contribution from the architecture and different features.

\subsection{Three Combinatorial Optimization Problems for Our Experiment Design}\label{subsec:CO}

 %\begin{figure*}[tb]
 	%\includegraphics[width=\linewidth]{figures/6iters}
 	%\caption{The example of QRF-GNN work on a toy Max-Cut instance. The Max-Cut problem involves partitioning a graph's nodes into two sets such that the number of edges between the sets is maximized. Because of the recurrent feature, QRF-GNN aims to reclassify each node to the opposite class of the majority of its nearest neighbors at each iteration, thereby optimizing the cut. Therefore, it is able to achieve the optimal solution in just a few iterations, while non-recurrent architectures (PI-GNN) need more than 20 iterations. 
 % }
 %	\label{fig:6iters}
 %\end{figure*}

In this paper focus on three CO problems defined on graphs $G=(V, E)$. It is worth noting that our approach can potentially be applied to any problem that can be formulated as QUBO. 
\\ \\
\textbf{Maximum Cut}.
The Max-Cut problem involves partitioning the vertices 
$V$ into two subsets such that the number of edges with endpoints in different subsets is maximized (or the total weight of such edges in the case of a weighted graph). Its QUBO formulation is as follows:
\begin{equation}
 \label{eq:qubo-maxcut}
  % \begin{split}
  \min_{x_i \in \left\{ 0, 1 \right\}} \mathcal{F}(x) = \sum_{i<j} A_{ij} (2 x_i x_j - x_i - x_j),\quad \forall i \in V
  % \end{split}
\end{equation}
\noindent where $A$ is an adjacency matrix of $G$. The decision variable $x_i = 1$ indicates that vertex $i$ is in one subset, while $x_i = 0$ indicates that it belongs to another one. In this paper we consider only unweighted graphs with $A_{ij}=1$,~$ \forall (i,j) \in E$. 

% The objective of \ref{eq:qubo-maxcut} can be reformulated:
%in a more physics-inspired \cite{Schuetz2022} way as
% \begin{equation}
%  \label{eq:qubo-maxcut-xQx}
%     \mathcal{F}(x) = \mathbf{x}^{T} Q \mathbf{x} = \sum_{i,j} x_i Q_{ij} x_j, \quad Q = A - D,
% \end{equation}
% \noindent where $D$ is a diagonal matrix with degrees of corresponding nodes. %Figure \ref{fig:6iters} shows the step-by-step work of QRF-GNN on the toy Max-Cut instance of 10 vertices and 12 edges, taking discretized solution vector at each iteration.

\textbf{Graph Coloring}.
We address two distinct versions of the graph coloring problem. The first one involves coloring a graph $G=(V, E)$ using a predetermined number of colors to minimize the number of adjacent vertices sharing the same color, thereby avoiding violations. The second formulation seeks to determine the minimum number of colors required to color the graph without violations. The QUBO formulations of both coloring problems is formulated as follows:
\begin{equation}
 \label{eq:qubo-coloring}
 \begin{split}
  & \min_{x_{i, c} \in \left\{ 0, 1 \right\}} \mathcal{F}(x) = \sum_i \left( 1 - \sum_c x_{i, c} \right) ^2 + \sum_{(i, j) \in E} \sum_c x_{i,c} x_{j, c}, \quad \forall i \in V,\  \forall c \in \{1,\dots,k\},
 \end{split}
\end{equation}
\noindent where $k$ is the number of colors the graph has to be colored.
For the second coloring problem $k$ is not fixed.
The loss function can be reduced to the second term of the objective \ref{eq:qubo-coloring} in order to train QRF-GNN or PI-GNN.
The condition for the uniqueness of the color assigned to the node which is specified by the first term is met automatically by the softmax layer.

\textbf{Maximum Independent Set}.

For a given graph $G=(V, E)$ the Maximum Independent Set (MIS) problem is to find a subset $S \subset V$ of pair-wise nonadjacent nodes of the maximum size $|S|$.
The QUBO formulation of MIS is as follows:
\begin{equation}
 \label{eq:qubo-mis}
  \begin{split}
  &\min_{x_{i} \in \left\{ 0, 1 \right\}} \mathcal{F}(x) = \quad - \sum_{i \in V }  {x_{i}} + P \sum_{ (i,j) \in E} {x_{i}x_{j}} \quad \forall i \in V.
  \end{split}
\end{equation}

where $x_{i} = 1$ if node $i \in S$ and $P$ is a penalty coefficient for violating the independence condition, which ensures that no two adjacent nodes are both included in the independent set.
Unlike the previously considered problems, this formulation contains a penalty term, raising the question of which value of $P$ should be chosen. In our algorithm an adaptive coefficient was applied. For more details, please see Appendix \ref{subsec:MIS-2}.
% In our experiments, we found that setting a small $P$ leads to the fact that the solutions found by the algorithm for a given number of iterations contain too many violations.
% A large $P$ value can cause the algorithm to quickly converge to a trivial solution with zero set size.
% To circumvent the problem of adjusting $P$ for different types of graphs, we propose in this paper to linearly increase the penalty value from $0.01$ to $2$ throughout all the iterations.
% This allows the algorithm to start the search in the space of large sets with violations while gradually narrowing the search space towards sets without violations.

%The described architectural choices of QRF-GNN lead to significant performance improvements, as is shown below.
%After all, the entire described approach of QRF-GNN leads to significant performance improvements, as will be seen below.

\section{Numerical experiments}
\label{sec:experiments}
In this section, we experimentally evaluated the performance of the algorithm on three popular combinatorial optimization problems, namely maximum cat, maximum independent set and graph coloring.
The benchmark data included both generated random graphs and well-known graphs of various structures of synthetic and real-world origin. We selected the best among learning algorithms and classical heuristics for each CO problem to compare the results. 

We found out that there was no need to do a time-consuming optimization of QRF-GNN hyperparameters (as it is done in the PI-GNN algorithm) to overcome existing GNNs.
However, the hyperparameters tuning may lead to better results of QRF-GNN.
The only parameters that we changed for different problems were the size of hidden layers and the maximum number of iterations, since for some types of graphs fewer iterations were sufficient to get a competitive result. As the solution to which the algorithm converges depends on the initialization, it is preferable to do multiple runs with different seeds to get the best result. We provide a detailed description of the experimental setup in Appendix \ref{sec:setup}.

% The computational time depended on the size of the graph and the achievement of convergence, but the longest run of QRF-GNN did not exceed 17 minutes.

% In all the following tables the best results are in bold and gaps mean data are not published.

\subsection{Maximum Cut}\label{subsec:maxcut}

To evaluate the performance of QRF-GNN, we compared results with other GNNs and popular heuristics on randomly generated regular graphs and the synthetic dataset Gset by \cite{Gset}.
\begin{wraptable}{l}{7.5cm}
% \begin{table}[b]
 \caption{\emph{P-value} of EO, PI-GNN, RUN-CSP and QRF-GNN for d-regular graphs with 500 nodes and different degree $d$ averaged over 200 randomly generated instances. }
\label{table:res-maxcut-dreg}
\begin{center}
\begin{small}
\begin{sc}
\begin{tabular}{l | l | l l l}
 \toprule
 \multirow{2}{*}{\bf d}  & \multicolumn{1}{c|}{heur}  & \multicolumn{3}{c}{Unsupervised Learning}  \\
\cmidrule{2-5}
&\multicolumn{1}{c|}{\bf EO}  &\multicolumn{1}{c}{\bf PI-GNN} &\multicolumn{1}{c}{\bf RUN-CSP} &\multicolumn{1}{c}{\bf QRF-GNN} \\
 
 \midrule
 3 & \textbf{0.727} & 0.612 & 0.714 & 0.725  \\
 5 & 0.737 & 0.608 & 0.726 & \textbf{0.738} \\
 10 & 0.735 & 0.658 & 0.710 & \textbf{0.737} \\
 15 & 0.736 & 0.644 & 0.697 & \textbf{0.739} \\
 20 & 0.732 & 0.640 & 0.685 & \textbf{0.735} \\
\bottomrule
\end{tabular}
\end{sc}
\end{small}
\end{center}
\vskip -0.1in
% \end{table}
\end{wraptable}
For the synthetic data the number of cut edges was counted as a metric.

For random regular graphs with the number of vertices $n \to \infty$, there exists a theoretical estimate of the maximum cut size that depends on $n$, vertex degree $d$ and a universal constant $P_* \simeq 0.7632$. Such asymptotics led to the appearance of a metric 
% motivated by the constant $P_*$
which is calculated as \emph{P-value}~$=\sqrt{\frac{4}{d}} \left(\frac{z}{n} - \frac{d}{4} \right)$, where $z$ corresponds to the obtained cut size (see \cite{Yao2019}). 

A higher \emph{P-value} corresponds to a better cut, and the closer it is to $P_*$, the nearer the solution found is to the best possible.

Table \ref{table:res-maxcut-dreg} shows the mean over 200 graphs \emph{P-value} obtained with specialized extremal optimization heuristics (EO) by \cite{Boettcher2003}, RUN-CSP and QRF-GNN for random regular graphs with $n=500$ nodes and different degrees $d$.
For QRF-GNN and PI-GNN we present results for the best value out of 5 runs. The PI-GNN architecture with GCN layers and setup as in \cite{Schuetz2022} was used for the calculations on graphs with $d>5$, other values are taken from the original paper.
Results of EO and RUN-CSP were presented by \cite{Tonshoff2021}, where RUN-CSP was allowed to make 64 runs for each graph.  QRF-GNN clearly outperforms RUN-CSP and PI-GNN in all cases and starting from $d=5$ shows the best results over all considered algorithms. Increasing the number of runs to 15 allows QRF-GNN to show the best result on graphs with $d=3$ with $\emph{P-value}=0.727$.
 
Table \ref{table:res-maxcut-gset} contains the comparison results on Gset of our algorithm with PI-GNN, RUN-CSP, EO and the SOTA heuristics Breakout Local Search (BLS) by \cite{BENLIC20131162} and the Tabu Search based Hybrid Evolutionary Algorithm (TSHEA) by \cite{WU2015827}. To obtain the data, the $\tau$-EO heuristic was implemented according to \cite{Boettcher2001} (details in \ref{subsec:Max-Cut_app} in the appendix).

QRF-GNN as well as BLS and TSHEA was run 20 times on each graph and the best attempt out of 64 was chosen for RUN-CSP. For the results of PI-GNN, the authors applied intensive hyperparameter optimization for each graph separately (see \cite{Schuetz2022}). Our algorithm outperforms both neural network approaches and the EO heuristic.  
Compared to the SOTA heuristics, QRF-GNN finds fewer cut edges on most small graphs, but on the largest presented graph G70 it not only yields the best number of cuts, but also requires only $\sim 1000$~s, while BLS and TSHEA took several times more, or about $\sim 11000$~s and $\sim 7000$~s correspondingly (see \cite{BENLIC20131162, WU2015827}).

\begin{table*}[h]
\caption{Number of cut edges for benchmark instances from Gset \cite{Gset} with the number of nodes $\vert V \vert$ and the number of edges $\vert E \vert$. QRF-GNN outperforms all other GNN-based approaches and the EO heuristic, while being comparable to SOTA heuristics.}
\label{table:res-maxcut-gset}
\vskip 0.15in
\begin{center}
\begin{small}
\begin{sc}
\begin{tabular}{l l l | l l l | l l l}
 \toprule
 \multirow{2}{*}{ Graph} & \multirow{2}{*}{$\vert V \vert$} & \multirow{2}{*}{ $\vert E \vert$} &
 \multicolumn{3}{c|}{heuristics}  & \multicolumn{3}{c}{ Unsupervised Learning}  \\
\cmidrule{4-9}
  & & & \multicolumn{1}{c}{\bf BLS} & \multicolumn{1}{c}{\bf TSHEA} & \multicolumn{1}{c|}{\bf EO} & \multicolumn{1}{c}{\bf PI-GNN} &
 \multicolumn{1}{c}{\bf RUN-CSP} & \multicolumn{1}{c}{\bf QRF-GNN} \\
  
\midrule
 G14 & 800 & 4694 & \textbf{3064}& \textbf{3064} & 3058 & 3026 & 2943 & 3058 \\
 G15 & 800 & 4661 & \textbf{3050}& \textbf{3050} & 3046 & 2990 & 2928 & 3049 \\
 G22 & 2000 & 19990 & \textbf{13359}& \textbf{13359} & 13323 & 13181 & 13028 & 13340 \\
 G49 & 3000 & 6000 & \textbf{6000}& \textbf{6000} & \textbf{6000} & 5918 & \textbf{6000} & \textbf{6000} \\
 G50 & 3000 & 6000 & \textbf{5880}& \textbf{5880} & 5878 & 5820 & \textbf{5880} & \textbf{5880} \\
 G55 & 5000 & 12468 & 10294 & \textbf{10299} & 10212 & 10138 & 10116 & 10282 \\
 G70 & 10000 & 9999 & 9541& 9548 & 9433 & 9421 & 9319 & \textbf{9559} \\
\bottomrule
\end{tabular}
\end{sc}
\end{small}
\end{center}
\vskip -0.1in
\end{table*}

%We distinguished two formulations in the graph coloring problem.
%
%The first formulation requires to color the graph $\gG=(\gV, \gE)$ with a given number of colors without violations.
%The second formulation is to find the minimum number of colors in which the graph can be colored by the algorithm.
%Here, coloring a graph means assigning a certain color to each vertex, and violation occurs when adjacent vertices have the same color.
%
%The QUBO model is formulated as follows:
%
%\begin{equation}
% \label{eq:qubo-coloring}
% \begin{split}
%  &\min \quad \sum_i \left( 1 - \sum_c x_{i, c} \right) ^2 + \sum_{(i, j) \in \vE} \sum_c x_{i,c} x_{j, c}, \\
%  &\text{ s.t.} \quad x_{i, c} \in \left\{ 0, 1 \right\}, \quad \forall i \in \gV, \quad \forall c \in \{1,\dots,k\},
% \end{split}
%\end{equation}
%
%\noindent where $k$ is the number of colors the graph has to be colored.
%
%The loss function can be reduced to the second term of the objective \ref{eq:qubo-coloring} in order to train QRF-GNN or PI-GNN.
%The condition for the uniqueness of the color assigned to the vertex which the first term specifies is met automatically by the softmax layer.

\subsection{Graph Coloring}\label{subsec:coloring}
We performed experiments on synthetic graphs from the COLOR dataset \cite{COLOR} and additional calculations for three real-world citation graphs Cora, Citeseer and Pubmed \cite{Li2022RethinkingGN}. While the number of vertices in synthetic graphs ranged from 25 and 561, citation graphs contained up to about 20 thousand vertices. The detailed specifications can be found in Appendix \ref{subsec:Coloring-2}. 

Results of graph coloring are compared with other GNN-based architectures, such as PI-GNN by \cite{schuetz2022coloring}, GNN-1N by \cite{wang2023graph}, GDN by \cite{Li2022RethinkingGN} and RUN-CSP by \cite{Tonshoff2021}, and the SOTA heuristics HybridEA by \cite{HybridEA}. The code\footnote{\url{http://rhydlewis.eu/gcol/}} which was used for the HybridEA estimation is based on \cite{bookHybridEA}. 
To evaluate the results of QRF-GNN, we did up to 10 runs for some graphs, although most of them required only one run.

Table \ref{table:res-coloring-conflicts} shows the best result of algorithms for coloring a graph with a chromatic number of colors. A violation occurs if there are adjacent vertices of the same color in the graph. It can be seen that QRF-GNN shows the best results on all instances, even outperforming the results of the SOTA heuristic on some graphs. 

Table \ref{table:res-coloring-chromatic} shows the number of colors that the algorithm needs to color the graph without violations. 
\begin{table}[h]
\parbox{.49\linewidth}{
 \caption{The number of violations when coloring the graph with chromatic number of colors by HybridEA (HEA) heuristics and GNN-based methods GNN-1N, PI-GNN, GDN and QRF-GNN for citation graphs and graphs from the COLOR dataset.}
\label{table:res-coloring-conflicts}
\begin{center}
\begin{small}
\begin{sc}
\scalebox{0.7}{
\begin{tabular}{l | l | l l l l}
 \toprule
\multirow{2}{*}{ Graph}
 & heur & \multicolumn{4}{c}{ Unsupervised Learning}  \\
\cmidrule{2-6}
& \multicolumn{1}{c|}{\bf HEA} & \multicolumn{1}{c}{\bf GNN-1N} &
 \multicolumn{1}{c}{\bf PI-GNN} & \multicolumn{1}{c}{\bf GDN} & \multicolumn{1}{c}{\bf QRF-GNN} \\
\midrule
 homer  & \textbf{0 }& \textbf{0} & \textbf{0} & \textbf{0} & \textbf{0} \\
 myciel6  & \textbf{0} & \textbf{0} & \textbf{0} & \textbf{0} & \textbf{0} \\
 queen5-5  & \textbf{0} & \textbf{0} & \textbf{0} & \textbf{0} & \textbf{0} \\
 queen6-6 & \textbf{0} & \textbf{0} & \textbf{0} & \textbf{0} & \textbf{0} \\
 queen7-7  & \textbf{0} & \textbf{0} & \textbf{0} & 9 & \textbf{0} \\
 queen8-8   & \textbf{0} & 1 & 1 & - & \textbf{0} \\
 queen9-9   & \textbf{0} & 1 & 1 & - & \textbf{0} \\
 queen8-12   & \textbf{0} & \textbf{0 }& \textbf{0} & \textbf{0} & \textbf{0} \\
 queen11-11  & 14 & 13 & 17 & 21 & \textbf{7} \\
 queen13-13  & 18 & 15 & 26 & 33 & \textbf{15} \\
\midrule
%& & & & & & & & \\
 Cora  & \textbf{0} & 1 & \textbf{0} & \textbf{0} & \textbf{0} \\
 Citeseer & \textbf{0} & \textbf{0} & \textbf{0} & \textbf{0} & \textbf{0} \\
 Pubmed  & \textbf{0} & - & 17 & 21 & \textbf{0} \\
\bottomrule
\end{tabular}}
\end{sc}
\end{small}
\end{center}
\vskip -0.1in
}
\hfill
\parbox{.49\linewidth}{
 \caption{The number of color needed for coloring without violations by HybridEA (HEA) heuristics and GNN-based methods PI-GNN, RUN-CSP and QRF-GNN on citation graphs and graphs from the COLOR dataset. Here $\chi$ is a known chromatic number.}
\label{table:res-coloring-chromatic}
\begin{center}
\begin{small}
\begin{sc}
  \scalebox{0.7}{
\begin{tabular}{l l | l | l l l}
  \toprule
\multirow{2}{*}{Graph} & \multirow{2}{*}{$\chi$} & heur & \multicolumn{3}{c}{ Unsupervised Learning}  \\
\cmidrule{3-6}
& & \multicolumn{1}{c|}{\textbf{HEA}}&
\multicolumn{1}{c}{\textbf{PI-GNN}} & \multicolumn{1}{c}{\textbf{RUN-CSP}} & \multicolumn{1}{c}{\textbf{QRF-GNN}} \\

\midrule
 homer & 13 & \textbf{13} & \textbf{13} & 17 & \textbf{13} \\ 
 myciel6 & 7 & \textbf{7} & \textbf{7} & 8 & \textbf{7} \\
 queen5-5 & 5 & \textbf{5} & \textbf{5} & \textbf{5} & \textbf{5} \\
 queen6-6 & 7 & \textbf{7} & \textbf{7} & 8 & \textbf{7} \\
 queen7-7 & 7 & \textbf{7} & \textbf{7} & 10 & \textbf{7} \\
 queen8-8 & 9 & \textbf{9} & 10 & 11 & \textbf{9} \\
 queen9-9 & 10 & \textbf{10} & 11 & 17 & \textbf{10} \\
 queen8-12 & 12 & \textbf{12} & \textbf{12} & 17 & \textbf{12} \\
 queen11-11 & 11 & \textbf{12} & 14 & $>$17 & \textbf{12} \\
 queen13-13 & 13 & \textbf{14} & 17 & $>$17 & 15 \\
\midrule
 Cora & 5 & \textbf{5} & \textbf{5} & - & \textbf{5} \\
 Citeseer & 6 & \textbf{6} & \textbf{6} & - & \textbf{6} \\
 Pumbed & 6 & \textbf{8} & 9 & - & \textbf{8} \\
\bottomrule
\end{tabular}}
\end{sc}
\end{small}
\end{center}
\vskip -0.1in
}
\end{table}
Since the authors of GNN-1N and GDN do not obtain results for this problem formulation in the original papers, we omitted them and compared additionally with RUN-CSP. For QRF-GNN, we successively increased the number of colors to find the optimal one.
In this problem formulation, QRF-GNN also surpasses existing GNN-based solutions, and is inferior to the HybridEA heuristics on only one graph.

\subsection{Maximum Independent Set}\label{subsec:mis}
%
%For a given graph $\gG=(V, E)$ the Maximum Independent Set (MIS) problem is to find a subset $S \subset V$ of pair-wise nonadjacent vertices of the maximum size $|S|$.
%The QUBO formulation of MIS is as follows:
%
%\begin{equation}
% \label{eq:qubo-mis}
%  \begin{split}
%  &\min \quad - \sum_{i \in V }  {x_{i}} + P \sum_{ (i,j) \in E} {x_{i}x_{j}}  \\
%  &\text{ s.t.} \quad x_{i} \in \left\{ 0, 1 \right\}, \quad \forall i \in V.
%  \end{split}
%\end{equation}
%
%where $x_{i} = 1$ if vertex $i \in S$ and $P$ is a penalty coefficient.

For MIS, we conduct experiments on randomly generated graphs of different structures.
The results shown in Table \ref{table:mis} represents a comparison of QRF-GNN with the range of cutting edge learning-based algorithms on two sets of Erdos-Renyi (ER) random graphs of 700-800 vertices and 9000-11000 vertices of 500 graphs each (see \cite{Erdos1984OnTE}). 
\begin{wraptable}{l}{7.5cm}
 \caption{Comparison of average found MIS sizes and runtime for QRF-GNN, learning-based methods and the SOTA heuristics KaMIS on sets of 500 Erdos-Renyi random graphs with a different number of nodes.}
\label{table:mis}
\begin{center}
\begin{sc}
\scalebox{0.7}{
\begin{tabular}{l l | c c | c c}
\toprule
 \multirow{2}{*}{Method} & \multirow{2}{*}{Type}  & \multicolumn{2}{c|}{ER-[700-800]}  & \multicolumn{2}{c}{ER-[9000-11000]} \\
\cmidrule{3-6}
  & &  Size & Time & Size & Time \\
\midrule
  KaMIS & Heur  & \textbf{44.87} & 52:13 & 374.57 & 7:37:21  \\
\midrule
 Intel & SL & 34.86 & 6:04 & 284.63 & 5:02 \\
 DIFUSCO & SL &  40.35 & 32:98 & - & -  \\
 T2TCO & SL  & 41.37 & 29:44 & - & - \\
 LwD & RL & 41.17 & 6:33 & 345.88 & 1:02:29 \\
 DIMES & RL & 42.06 & 12:01 & 332.8 & 12:31\\
 GFlowNets & UL & 41.14 & 2:55 & 349.42 & 1:49:43 \\
\midrule
 QRF-GNN & UL & 42.45 & 3:46 & \textbf{375.44} & 10:32\\
\bottomrule
\end{tabular}
}
\end{sc}
\end{center}
\vskip -0.1in
\end{wraptable}
For the comparison, we selected SOTA Supervised Learning (SL) methods: Intel by \cite{Li2018CombinatorialOW}, Difusco by \cite{sun2023difusco} and T2TCO by \cite{li2023from}; LwD by \cite{Ahn} and DIMES by \cite{qiu2022dimes} are recent Reinforement Learning (RL) methods and GflowNets by \cite{zhang2023let} is Unsupervised (UL) applying Generative Flow Networks for CO.
We also consider the results of the SOTA MIS solver Kamis (ReduMIS) \cite{Lamm2017}. For some of the baselines there are different published results related to the various (greedy or sampling) solution decoding, and we take the results of the best average MIS size.
Since Difusco and T2TCO are not scalable to solve large problems from the ER-9000-11000 dataset (see \cite{zhang2023let}), we omitted their results. 
As can be seen from Table \ref{table:mis}, QRF-GNN performs the best among all learning methods.
This is especially evident on large graphs, where it surpasses the results of Kamis, being significantly superior in terms of speed.

We also follow \cite{Tonshoff2021} and make experiments on a collection of hard instances with hidden optimal solutions generated by the RB model \cite{XU2006291} (see Table \ref{table:mis1}). In this case, QRF-GNN outperforms both RUN-CSP and greedy baselines, and is slightly inferior to KaMIS.

\begin{table}[h]
 \caption{MIS results of QRF-GNN, GNN-based method RUN-CSP, greedy and SOTA KaMIS heuristics on RB Model graphs. We report average MIS sizes over 5 runs with standard deviations.}
\label{table:mis1}
\begin{center}
\begin{sc}
\scalebox{0.85}{
\begin{tabular}{l l l | c c | c c}
\toprule
 \multirow{2}{*}{Graph} & \multirow{2}{*}{$ \vert V \vert$ }  & \multirow{2}{*}{ $\vert E \vert $} & \multicolumn{2}{c|}{Heuristics}  & \multicolumn{2}{c}{Unsupervised Learning} \\
\cmidrule{4-7}
  &  & & \multicolumn{1}{c}{ KaMIS} &
\multicolumn{1}{c|}{Greedy} & \multicolumn{1}{c}{ RUN-CSP} & \multicolumn{1}{c}{ QRF-GNN}\\
\midrule
 frb30-15 & 450 & 18k & 30 $\pm$ 0.0 & 24.6 $\pm$0.5  &  25.8 $\pm$ 0.8  & 28.4 $\pm$ 0.4\\
 frb40-19 & 790 & 41k & 39.4 $\pm$ 0.5 & 33.0 $\pm$1.2  & 33.6 $\pm$ 0.5 &  36.8 $\pm$ 0.7\\
 frb50-23 & 1150 & 80k & 48.8 $\pm$ 0.4 & 42.2 $\pm$0.8 & 42.2 $\pm$ 0.4 &  45 $\pm$ 0.6\\
 frb59-26 & 1478 & 126k & 57.4 $\pm$ 0.9 & 48.0 $\pm$ 0.7 & 49.4 $\pm$ 0.5 & 54.6 $\pm$ 1\\
\bottomrule
\end{tabular}
}
\end{sc}
\end{center}
\vskip -0.1in
\end{table}

\section{Impact of the recurrence connection on GNNs performance.}
\label{sec:ablation_main}
\begin{figure*}[h]
	\centering
	\includegraphics[width=0.95\linewidth]{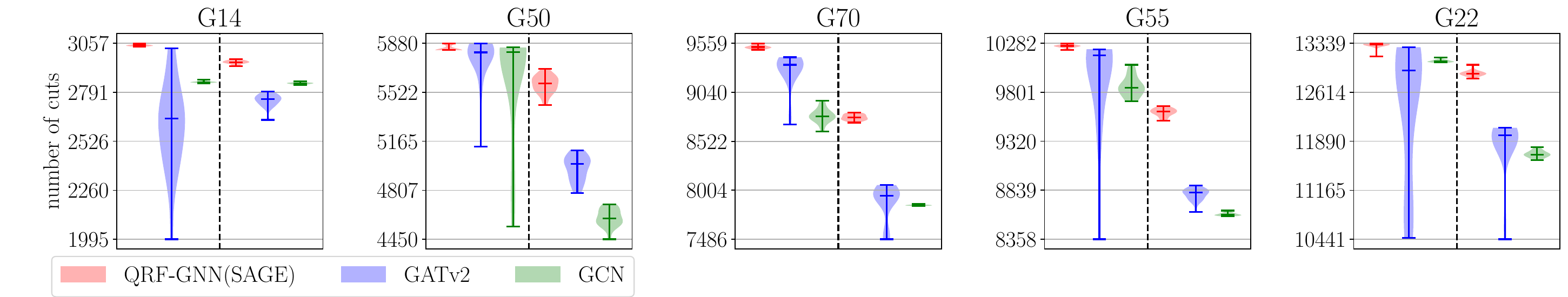}
	\caption{Distribution of the results over 20 runs of GNNs with different convolutional layers with (left) and without (right) recurrent connection on several instances from Gset.
	% Default means the QRF-GNN architecture results. 
 Horizontal lines mean maximum, median and minimum values.
	}
	\label{fig:all_rec_no_rec}
\end{figure*}

We have investigated how the recurrent use of vertex class membership affects architectures with different types of convolutions,
namely GCN \cite{kipf2017semisupervised} and GATv2 \cite{brody2021attentive} in addition to the default architecture of QRF-GNN with SAGE convolutions.
For this purpose, we constructed GNNs from two consecutive convolutions of a given type.
The sizes of hidden states were the same for all networks and coincided with the sizes of the described default QRF-GNN architecture.
We performed 20 calculations with different seeds for graphs from Gset to solve the Max-Cut problem.
Figure \ref{fig:all_rec_no_rec} shows that the type of recurrence proposed in this paper greatly improves the maximum cut found for all types of convolutions.
At the same time, the default QRF-GNN shows better results compared to other architectures. 

% In the case of GATv2 convolutional layer, this result can be explained by the fact that the QUBO formulation does not imply differences in the importance of one neighbor over another, while this type of convolution includes the use of attention score between nodes.

\section{Related work}
\label{sec:gnn}
%The PI-GNN approach does not involve the use of a labeled data set.

Graph neural networks are rapidly gaining popularity as a powerful tool for solving CO problems \cite{Cappart23}.
Supervised learning based approaches are commonly applied \cite{Prates2019,Chen2018,GasseCFCL19,Li2018CombinatorialOW,sun2023difusco,li2023from}. 
However, the need to collect labeled training instances into representative and unbiased dataset is a significant limitation of supervised algorithms. Additionally, they often face challenges with generalization to new, unseen problem instances.
Reinforcement learning (RL) presents an alternative by generating iterative solutions \cite{MAZYAVKINA2021105400,KhalilDZDS17, kool2018attention,qiu2022dimes}. While being promising approach, reinforcement learning methods may experience difficulties when facing large scale problems due to the vastness of the state space, and the need of a large number of samplings.
The unsupervised learning paradigm, where solvers do not require a training set of pre-solved problems, has the potential to overcome these limitations. \cite{Tonshoff2021} proposed RUN-CSP as a recurrent GNN to solve maximum constraint satisfaction. \cite{Amizadeh2018} developed GNN to solve SAT and CircuitSAT. \cite{Karalias2020} train GNN to obtain a distribution of nodes corresponding to the candidate solution and \cite{sun2023annealed} provided an
annealed version of it. \cite{wang2022unsupervised} study entry-wise concave relaxations of CO objectives. \cite{Schuetz2022} apply relaxed QUBO as instance specific GNN loss, \cite{schuetz2022coloring, wang2023graph} extend it for solving graph coloring problem.

%The base  architecture of PI-GNN consists of a trainable embedding layer to produce input features of nodes and
%several graph convolutional layers (GCN by \cite{kipf2017semisupervised} or GraphSAGE by \cite{hamilton2017sage})
%\cite{ichikawa2024controlling} suggested an improvement based on incorporating annealing and regularising terms.
%\cite{wang2023unsupervised}

%In recent work \cite{wang2023graph} authors have introduced GNN-1N, adapting negative message passing technique into unsupervised GNN for solving graph coloring problem.

%The ability of unsupervised GNNs, and in particular those using the relaxed QUBO as a loss function, to accurately solve CO problems has been debated in the scientific community.
%Responding to ~\cite{Schuetz2022}, ~\cite{Boettcher2023} had made a comparative analysis for the Max-Cut on 3-regular graphs and concluded that GNN based approach can not outperform greedy algorithms and provided solutions
%much worse than the reputable EO ~\cite{Boettcher2001} heuristics.
%In ~\cite{Angelini2023} authors also claim that modern GNNs do worse than classical greedy algorithms in solving CO problems, providing experiments on random instances for MIS.

%In our work, we show a rather simple but elegant method how to overcome those limitations of GNNs and successfully improve performance while retaining the quality of suboptimal solutions.

\section{Conclusion}
\label{sec:conclusions}
In this work, we propose the novel recurrent algorithm QRF-GNN for graph neural networks, designed for solving combinatorial optimization problems in unsupervised mode. We show that our novel recurrent feature update method significantly enhances the performance of all types of GNN convolutions considered. 
We conduct computational experiments on the well-known maximum cut, graph coloring and maximum independent set problems. The results of our comparative analysis demonstrate that QRF-GNN drastically outperforms all learning-based baselines, including SOTA supervised, unsupervised, and reinforcement learning methods. Moreover, we show that QRF-GNN competes with the best classical heuristics for the problems addressed while showing a distinct advantage in computational time on large graphs. 
% For example, on large Erdos-Renyi random graphs, it outperforms KaMIS, being almost 50 times faster.

For the future work, we consider QRF-GNN superior performance and scalability promising to be extended to other CO problem formulated as QUBO, thus highlighting its potential in the field of combinatorial optimization.

\newpage
\bibliographystyle{unsrtnat} % Style
\bibliography{pr-gnn.bib}

%\section{Reproducibility Statement}
%\label{sec:reproducibility}
%\input{reproducibility.tex}
%\section{Impact Statement}
%This paper presents work whose goal is to advance the field of Machine Learning. There are many potential societal consequences of our work, none which we feel must be specifically highlighted here.
%\printbibliography
%\bibliography{example_paper}

%\bibliography{pr-gnn}
%\bibliographystyle{icml2024}

%%%%%%%%%%%%%%%%%%%%%%%%%%%%%%%%%%%%%%%%%%%%%%%%%%%%%%%%%%%%%%%%%%%%%%%%%%%%%%%
%%%%%%%%%%%%%%%%%%%%%%%%%%%%%%%%%%%%%%%%%%%%%%%%%%%%%%%%%%%%%%%%%%%%%%%%%%%%%%%
% APPENDIX
%%%%%%%%%%%%%%%%%%%%%%%%%%%%%%%%%%%%%%%%%%%%%%%%%%%%%%%%%%%%%%%%%%%%%%%%%%%%%%%
%%%%%%%%%%%%%%%%%%%%%%%%%%%%%%%%%%%%%%%%%%%%%%%%%%%%%%%%%%%%%%%%%%%%%%%%%%%%%%%

%%%%%%%%%%%%%%%%%%%%%%%%%%%%%%%%%%%%%%%%%%%%%%%%%%%%%%%%%%%%

\newpage
\appendix
\onecolumn
%\section{You \emph{can} have an appendix here.}
%
%You can have as much text here as you want. The main body must be at most $8$ pages long.
%For the final version, one more page can be added.
%If you want, you can use an appendix like this one.
%
%The $\mathtt{\backslash onecolumn}$ command above can be kept in place if you prefer a one-column appendix, or can be removed if you prefer a two-column appendix.  Apart from this possible change, the style (font size, spacing, margins, page numbering, etc.) should be kept the same as the main body.
%%%%%%%%%%%%%%%%%%%%%%%%%%%%%%%%%%%%%%%%%%%%%%%%%%%%%%%%%%%%%%%%%%%%%%%%%%%%%%%
%%%%%%%%%%%%%%%%%%%%%%%%%%%%%%%%%%%%%%%%%%%%%%%%%%%%%%%%%%%%%%%%%%%%%%%%%%%%%%%
\section{Technical Details and Convergence}
\label{sec:training}
\subsection{General experimental setup}
\label{sec:setup}

All random graphs in this work were generated by the NetworkX\footnote{\url{https://networkx.org/}} package.
To implement the QRF-GNN architecture the DGL library\footnote{\url{https://www.dgl.ai/}} was used. 
The pseudocode of one iteration with the QRF-GNN architecture is presented in the Algorithm \ref{alg:forward_prgnn}.
We used the Adam optimizer without a learning rate schedule. The learning rate was set empirically to 0.014, the rest of the optimizer parameters remained set by default.
Gradients were clipped at values 2 of the Euclidean norm.

We limited the number of iterations to $5 \times 10^4$ for random regular graphs and $10^5$ for all the other graphs, but in some cases convergence was reached much earlier.
If the value of the loss function at the last 500 iterations had differed by less than $10^{-5}$ it was decided that the convergence was achieved and the training was stopped.
In the case of the graph coloring problem, an additional stopping criterion was used and the solution was considered to be found when the absolute value of the loss function becomes less than $10^{-3}$.

The dropout was set to 0.5.
The dimension of the random part of input vectors was equal to 10, the size of hidden layers was fixed at 50 for Max-Cut and at 140 for graph coloring.

Due to the stochasticity of the algorithm, it is preferable to do multiple runs with different seeds to find the best result.
One can do separate runs in parallel possibly utilizing several GPUs.
If the device has enough memory, the RUN-CSP scheme by \cite{Tonshoff2021} can be used.
In this case, one composite graph with duplicates of the original one is created for the input.
We trained the model in parallel on the NVIDIA Tesla V100 GPU.
Conventional heuristics were launched on the machine with two Intel Xeon E5-2670 v3 @ 2.30GHz.

We tested three ways to recursively utilize the probability data.
Specifically, we passed raw probability data taken before the sigmoid layer, data after the sigmoid layer or concatenated both of these options.
Different recurrent features led to a minor improvement on some graphs, while at the same time slightly worsening the results on other graphs.
In this work we presented results for the concatenated data.
\begin{algorithm}[tbh]
    \caption{Forward propagation of the QRF-GNN algorithm at iteration $t$}\label{alg:forward_prgnn}
    \begin{algorithmic}[1]
    \STATE{\bfseries Input:} Graph $G(V, E)$, static nodes features $\{ a_i, \forall i \in V \}$
    \STATE{\bfseries Output:} Probability $p_i$,  hidden state $h^t_i, \forall i \in V$ \quad
    \STATE $h^{t, 0}_i \gets  \big[ a_i \ h^{t - 1}_i \big], \quad \forall i \in V$\;
        \FOR{$i \in V$}
        \STATE  $h^{t,1}_{N(i)} \gets \rho_{\text{mean}}  \left( \left\{ h^{t,0}_j, \forall j \in N(i) \right\} \right)$\;
        \STATE  $h^{t,1}_i \gets f \left( W^1 \big[h^{t,0}_i \ h^{t,1}_{N(i)}\big] \right)$\;

        \STATE  $h^{t,2}_{N(i)} \gets \rho_{\text{pool}}  \left( \left\{ h^{t,0}_j, \forall j \in N(i) \right\} \right)$\;
        \STATE  $h^{t,2}_i \gets f \left( W^2 \big[h^{t,0}_i \ h^{t,2}_{N(i)}\big] \right)$\;
        \ENDFOR

        \STATE  $\{ h^{t,1}_i\} \gets \text{BN}_{\gamma1, \beta1}(\{ h^{t,1}_i, \forall i \in V \})$\;
        \STATE  $\{ h^{t,2}_i \} \gets \text{BN}_{\gamma2, \beta2}(\{ h^{t,2}_i, \forall i \in V \})$\;
        \FOR{$i \in V$}
         \STATE $h^{t,12}_i \gets f(h^{t,1}_i + h^{t,2}_i)$\;

         \STATE $h^{t,12}_i \gets \text{Dropout}(h^{t,12}_i)$\;

         \STATE $h^{t,\text{out}}_{N(i)} \gets \rho_{\text{mean}}  \left( \left\{ h^{t,12}_j, \forall j \in N(i) \right\} \right)$\;
         \STATE $h^{t,\text{out}}_i \gets f \left( W^{\text{out}} \big[h^{t,12}_i \ h^{t,\text{out}}_{N(i)}\big] \right)$\;
        \ENDFOR
    \STATE$p_i, h^t_i \gets \sigma(h^{t,\text{out}}_i), h^{t,\text{out}}_i, \quad \forall i \in V$
\end{algorithmic}
\end{algorithm}

\subsection{Convergence}
\label{sec:convergence}

\begin{figure}[!ht]
    \begin{center}
    \begin{tabular}{cc}
	a \includegraphics[width=0.43\linewidth]{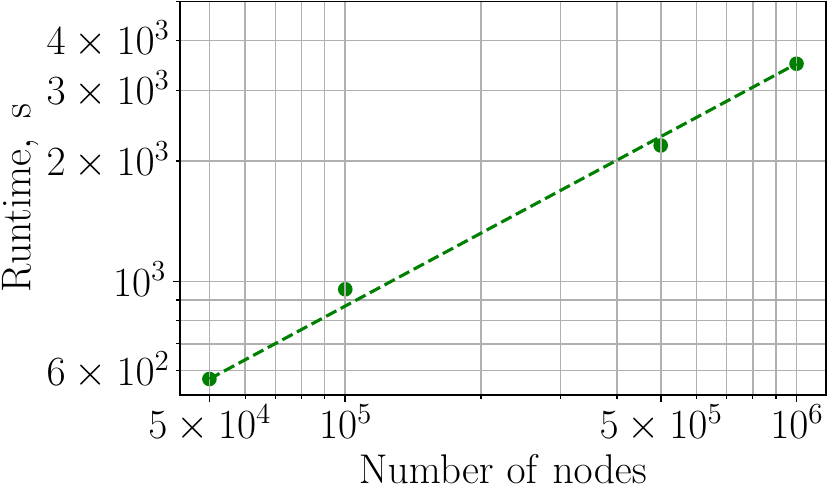} &
    b \includegraphics[width=0.49\linewidth]{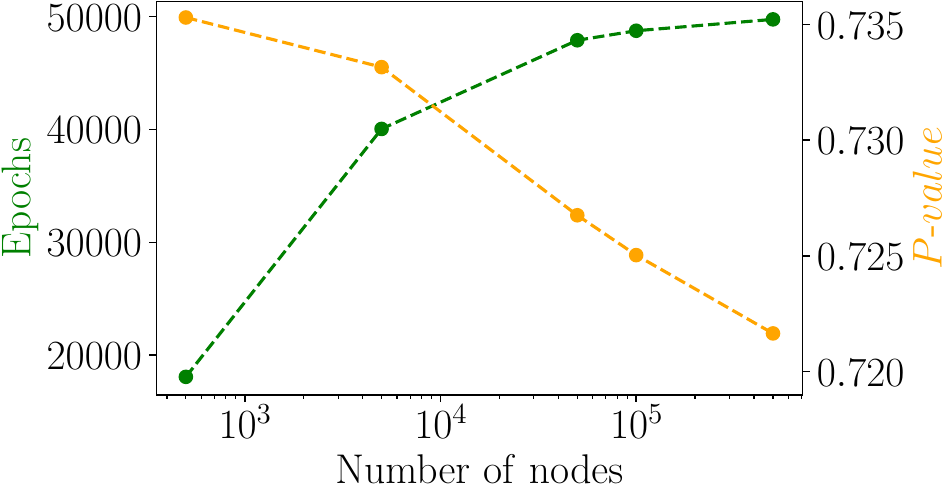}
	
    \end{tabular}
    \end{center}
    \caption{a) The computation time of $5\times 10^4$ iterations of QRF-GNN on random regular graphs with $d=5$ in the sparse format depending on the number of nodes.
    b) The iteration number averaged over 20 graphs at which the algorithm found the best solution during the training process (green) and
	the mean \textit{P-value} for 1 run (orange).}
	\label{fig:linear_scale}
\end{figure}

The number of runs and iterations in experiments were not optimal and were chosen for a more fair comparison with other algorithms.
More runs and iterations can lead to better results.
We conducted additional experiments on 20 random regular graphs with $d=5$ and up to one million nodes.
One run was made for each graph and the number of iterations was limited to $5 \times 10^4$.
The training time for large graphs in sparse format on single GPU is shown in Figure \ref{fig:linear_scale}a.
We also analyzed how the number of iterations can affect the quality of the solution.
As the number of vertices increases, the iteration at which the last found best solution was saved moves closer to the specified boundary (see Figure \ref{fig:linear_scale}b).
Meanwhile, the average \textit{P-value} of one run drops from $0.735$ to $0.722$ and one of the reasons for this may include the limited duration of training.
If, for example, we train QRF-GNN for $10^5$ iterations on graphs with $n=5\times 10^4$ nodes, the average \textit{P-value} will increase from $0.726$ to $0.728$,
while on small graphs with $n=500$ we do not observe such an effect.
Thus, it is difficult to talk about the convergence of the algorithm on large instances under the given constraint.
The recommendation is to follow the latest best solution updates and terminate the algorithm if it does not change for a sufficiently large ($>10^4$) number of iterations.

%\begin{figure}[tb]
%    \begin{center}
%    \includegraphics[width=0.6\linewidth]{figures/QRFGNN_On}
%	\caption{The mean runtime of QRF-GNN over 100 random regular graphs with $d=5$ depending on number of nodes.
%    Blue dots correspond to the case with a limit of $5 \times 10^4$ iterations per run.
%    Orange line and dots show the runtime required for the mean \textit{P-value} to be greater than $0.73$.
%	}
%	\label{fig:linear_scale}
%    \end{center}
%\end{figure}
%

\begin{figure}[tb]
	\centering
	\includegraphics[width=0.8\linewidth]{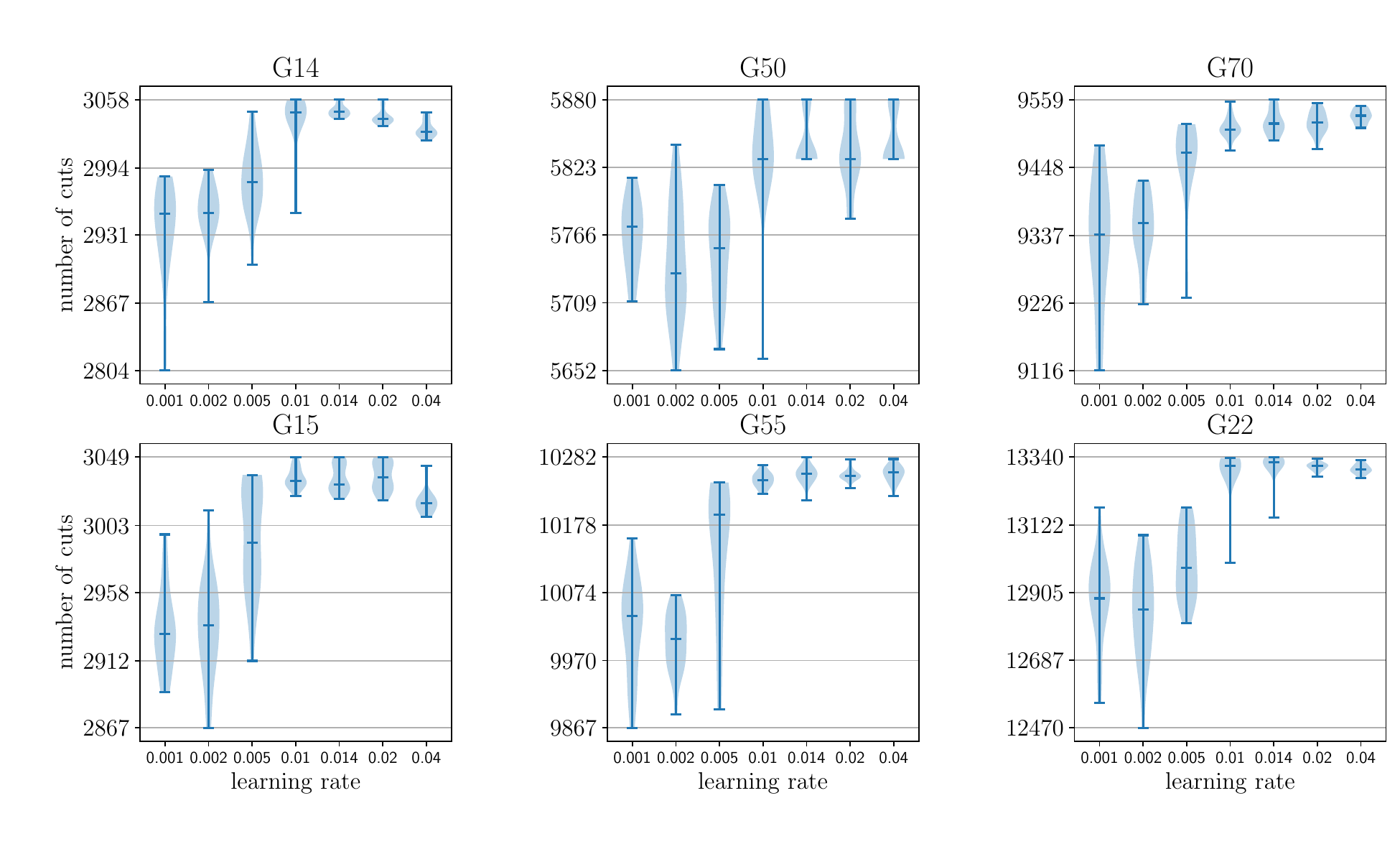}
	\caption{Results distribution for 20 runs of the default QRF-GNN architecture depending on the learning rate for the Max-Cut problem on several instances from Gset.
    The number of iterations in all cases was fixed at $10^5$.
	Horizontal lines mean maximum, median and minimum values.
	}
	\label{fig:diff_lr}
\end{figure}

In order to study the robustness of the algorithm with respect to changes in hyperparameters, we run the default QRF-GNN
architecture with two parallel layers on graphs from the Gset dataset for the Max-Cut problem.
All hyperparameters except the learning rate were chosen as described in Section \ref{sec:experiments}.
As can be seen from Figure \ref{fig:diff_lr}, small values of the learning rate do not allow to achieve convergence in $10^5$ iterations.
For the learning rate greater than $0.01$ the results become relatively stable and the best number of cuts is achieved for values from $0.01$ to $0.02$.

\subsection{Max-Cut}
\label{subsec:Max-Cut_app}

In Table \ref{table:res-maxcut-dreg} results of EO and RUN-CSP were taken from \cite{Tonshoff2021}, where \emph{P-value} was averaged over 1000 graphs.
RUN-CSP was allowed to make 64 runs for each graph and in the case of EO the best of two runs was chosen \cite{Yao2019}.
\emph{P-values} of PI-GNN depend on the particular architecture.
Results for graphs with a degree 3 and 5 were published in \cite{Schuetz2022} for the architecture with GCN layer, and it corresponds to the value in the column for PI-GNN.
The cut size was bootstrap-averaged over 20 random graph instances and PI-GNN took up to 5 shots.
In the paper by \cite{Schuetz2023_rep1} the authors considered another option with the SAGE layer and showed that in this case the results for graphs with a degree 3 can be improved by 10.78\%.
However, we did not notice an improvement over the GCN architecture on graphs with a higher degree.

To make the evaluation more informative, we implemented $\tau$-EO heuristic from~\cite{Boettcher2001}.
As suggested by authors, we set $\tau = 1.3$ and the number of single spin updates was limited by $10^7$.
For small graphs $\tau$-EO can find a high-quality solution, but with increase of the graph size the accuracy of the algorithm degrades due to the limited number of updates.
This behavior is expected by the authors, who suggested optimal scaling for number of updates as ${\sim}O(\vert V \vert^3)$.
However, it is computationally expensive to carry out the required number of iterations.
We performed 20 runs of EO with different initializations to partially compensate for this.
Within the given limit, the EO algorithm took $\sim$6800 seconds per run to obtain a solution for the relatively large graph G70.

QRF-GNN as well as BLS and TSHEA was run 20 times on each graph.
The best attempt out of 64 was chosen for RUN-CSP in original papers.
In order to obtain the results of PI-GNN, the authors applied hyperparameter optimization for each graph.
The results of RUN-CSP for the G70 graph was obtained by running the code\footnote{\url{https://github.com/toenshoff/RUN-CSP}} with parameters reported in \cite{Tonshoff2021}.

\subsection{Coloring}\label{subsec:Coloring-2}

To find the number of colors required to color the graph without violations, we successively increased the number of colors in each new run until the correct coloring was found among 10 seeds.
The number of nodes and edges of the investigated graphs is presented in Table \ref{table:ev}.
To evaluate the results of QRF-GNN, we did up to 10 runs for some graphs, although most of them required only one run.
The convergence time for PI-GNN and QRF-GNN is shown in Table \ref{table:res-coloring-runtime}.

Results for graphs anna, david, games120, muciel5, huck and jean were omitted in the main tables because all algorithms find optimal solutions without violations.
Results of GNN-1N for citation graphs were obtained by implementing the algorithm from the original paper. Since there was no instruction how to optimize hyperparameters, we took them close to PI-GNN and chose the best among 10 runs.

The number of iterations for convergence of QRF-GNN on citation graphs was no more than 6000 and varied for the COLOR dataset from $\sim$200 to $9 \times 10^4$.
The estimated runtime for QRF-GNN turned out to be significantly less than for PI-GNN, and in some cases the difference reaches more than three orders of magnitude.
This is due to the fact that QRF-GNN does not require exhaustive tuning of hyperparameters for each instance in contrast to PI-GNN \cite{schuetz2022coloring}.

\begin{table}[b]
 \caption{Approximate runtime in seconds for PI-GNN and QRF-GNN training on a single GPU on instances from the COLOR dataset and citation graphs.}
\label{table:res-coloring-runtime}
\vskip 0.15in
\begin{center}
\begin{small}
\begin{sc}
\begin{tabular}{l l l | l l}
\multicolumn{1}{c}{$\bf Graph$} & \multicolumn{1}{c}{$\vert V \vert$} & \multicolumn{1}{c|}{$ \vert E \vert$} &
\multicolumn{1}{c}{$\bf PI-GNN, \times 10^3 s$} & \multicolumn{1}{c}{$\bf QRF-GNN, \times 10^3 s$} \\
 \midrule

 COLOR & 25-561 & 160-3328 & 3.6 $\div$ 28.8 & 0.002 $ \div 1$ \\
 \midrule

 Cora & 2708 & 5429 & 0.3 & 0.06 \\
 Citeseer & 3327 & 4732 & 2.4 & 0.018 \\
 Pubmed & 19717 & 44338 & 24 & 0.156 \\
\end{tabular}
\end{sc}
\end{small}
\end{center}
\vskip -0.1in
\end{table}

\begin{table}[t]
 \caption{Number of Vertices and Edges in coloring graphs.}
\label{table:ev}
\vskip 0.15in
\begin{center}
\begin{small}
\begin{sc}

\begin{tabular}{l l l }
 \toprule
\multirow{1}{*}{$\bf Graph$} & \multirow{1}{*}{ $\vert V \vert$} & \multirow{1}{*}{$ \vert E \vert$} \\
\midrule
 anna & 138 & 493  \\
 david & 87 & 406  \\
 games120 & 120 & 638\\
 homer & 561 & 1629 \\
 huck & 74 & 301   \\
 jean & 80 & 254  \\
 myciel5 & 47 & 236 \\
 myciel6 & 95 & 755  \\
 queen5-5 & 25 & 160 \\
 queen6-6 & 36 & 290 \\
 queen7-7 & 49 & 476  \\
 queen8-8 & 64 & 728  \\
 queen9-9 & 81 & 1056   \\
 queen8-12 & 96 & 1368  \\
 queen11-11 & 121 & 1980  \\
 queen13-13 & 169 & 3328 \\
\midrule
%& & & & & & & & \\
 Cora & 2708 & 5429  \\
 Citeseer & 3327 & 4732  \\
 Pubmed & 19717 & 44338  \\
\bottomrule
\end{tabular}
\end{sc}
\end{small}
\end{center}
\vskip -0.1in
\end{table}

\subsection{MIS} \label{subsec:MIS-2}

In our experiments, we found that setting a small $P$ leads to the fact that the solutions found by the algorithm for a given number of iterations contain too many violations.
A large $P$ value can cause the algorithm to quickly converge to a trivial solution with zero set size.
To circumvent the problem of adjusting $P$ for different types of graphs, we propose in this paper to linearly increase the penalty value from $0.01$ to $2$ throughout all the iterations.
This allows the algorithm to start the search in the space of large sets with violations while gradually narrowing the search space towards sets without violations.

\subsection{Toy example of a Max-Cut problem}\label{subsec:toy}

We provide a toy example of a Max-Cut problem to illustrate the performance of our proposed QRF-GNN method compared to the PI-GNN approach. The problem instance consists of a graph with 12 edges and 10 nodes, as shown in Figure \ref{fig:6iters}.

An experimental setup for the QRF-GNN architecture is set by default similar to the description in section \ref{sec:experiments}. The PI-GNN architecture consists of a trainable embedding layer and one hidden layer, the sizes of which are set similarly to QRF-GNN and equal to 50.

For the considered problem, QRF-GNN finds the optimal solution (cut = 12) in an average of 10.96 iterations, while PI-GNN requires 532.1 iterations on average. These results are based on 100 runs with different random seeds.

% \begin{table}[h]
% \caption{Comparison of QRF-GNN and PI-GNN on a toy Max-Cut problem with 12 edges and 10 nodes. Results are averaged over 100 runs with different random seeds.}
% \centering
% \begin{small}
% \begin{sc}
% \begin{tabular}{l l }

% \multirow{1}{*}{\textbf{Algorithm}} & \multirow{1}{*}{\textbf{Average Number of Iterations to Optimal Solution}} \\
% \midrule
%  QRG-GNN & 10.96  \\
%  PI-GNN & 532.1  \\

% \end{tabular}
% \end{sc}
% \end{small}

% \label{table:maxcut_comparison}
% \end{table}

\newpage
\section{Ablation for QRF-GNN Components}
\label{sec:ablation}
\begin{figure}[t]
	\centering
	\includegraphics[width=0.8\linewidth]{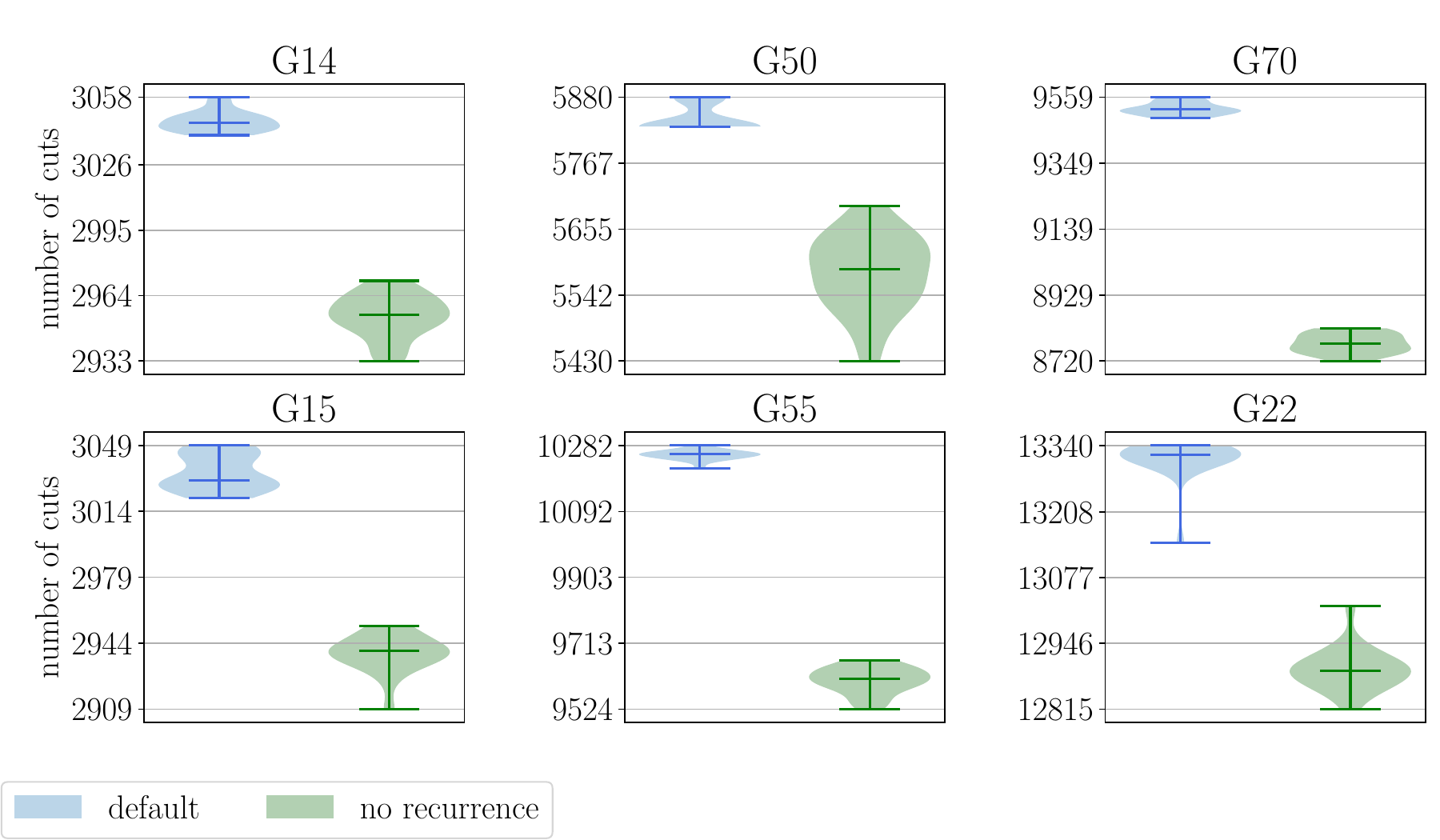}
	\caption{Results distribution for 20 runs of QRF-GNN with (blue) and without (green) the recurrent connection on several instances from Gset.
	Horizontal lines mean maximum, median and minimum values.
	}
	\label{fig:no_rec}
\end{figure}

We analyzed which components of QRF-GNN make the greatest contribution to its performance on the example of Max-Cut problem-solving.
The default architecture includes two intermediate SAGEConv layers and the recurrent feature.
The most dramatic drop in quality occurs if the recurrent part is excluded (see Figure \ref{fig:no_rec}).

\begin{figure}[t]
	\centering
	\includegraphics[width=0.8\linewidth]{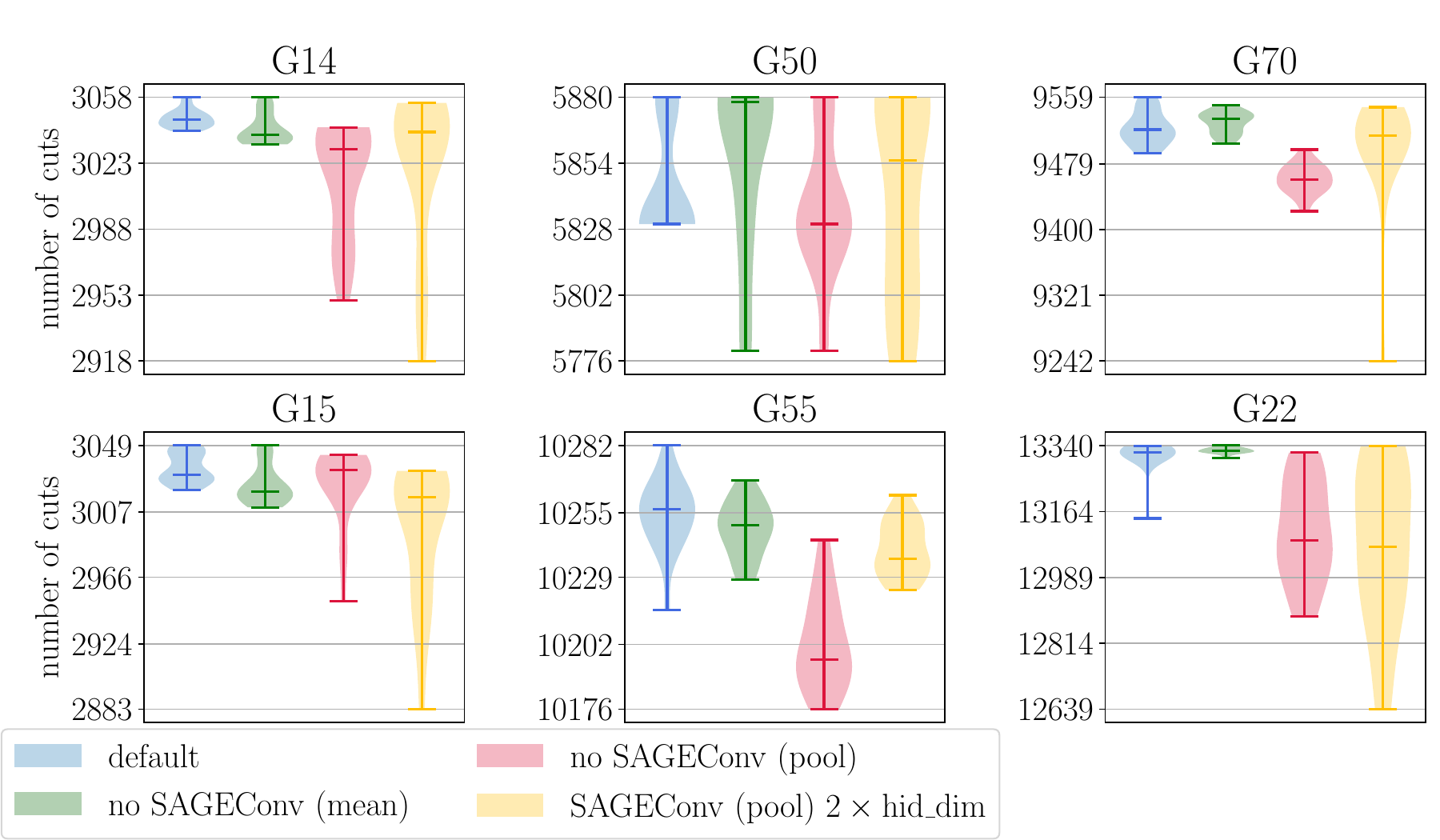}
	\caption{Results distribution for 20 runs of QRF-GNN of the default architecture (blue), with the absence of one SAGE layer with a mean aggregation function (green) or the SAGE layer with a pool aggregation function (red).
	Horizontal lines mean maximum, median and minimum values.
	}
	\label{fig:layers}
\end{figure}

\begin{figure}[t]
	\centering
	\includegraphics[width=0.6\linewidth]{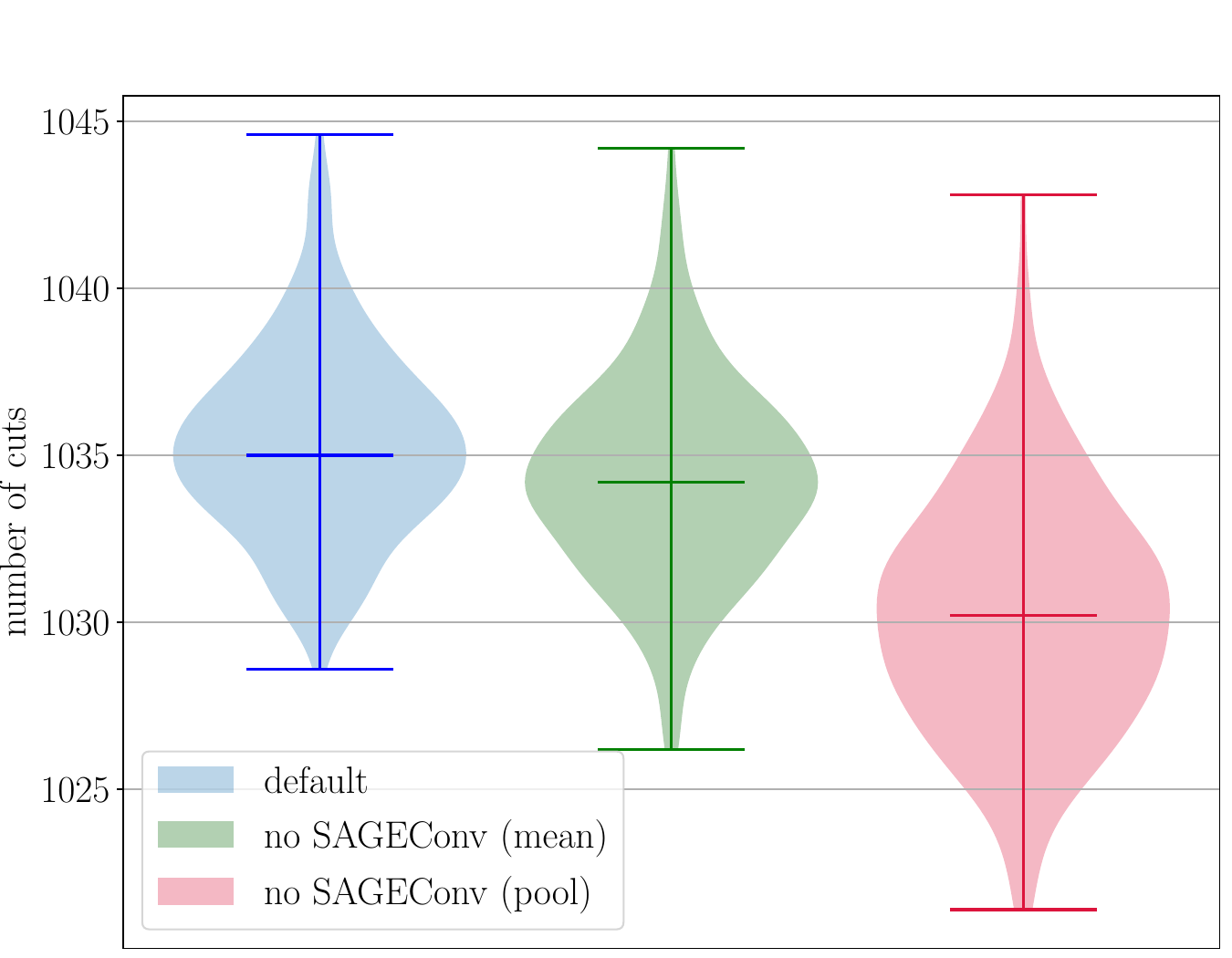}
	\caption{Results distribution for the mean of 5 runs on 200 random regular graphs with $d=5$ and 500 nodes.
	QRF-GNN had the same configurations as in the Fig. \ref{fig:layers}.
	}
	\label{fig:d5-layers}
\end{figure}

Throwing out one of the intermediate convolutional layer does not result in such a strong downgrade (see Figure \ref{fig:layers}).
However, the absence of the convolutional layer with a pool aggregation function leads to a decrease in the median result, upper bound, and an increase in the results dispersion for almost all graphs.
Discarding the layer with a mean aggregation function can increase the median cut and even decrease the variance in some cases, but upper bounds either stay the same or decrease,
even if we double the number of parameters in the remaining hidden layer (see Figure \ref{fig:layers}).
Further ablation on random regular graphs shows that the absence of any convolutional layer leads to a worse result(see Figure \ref{fig:d5-layers}).
Table \ref{table:ablation-maxcut-dreg} with average results for one seed and the best of the five seeds also confirms the advantage of using a combination of two layers.

\begin{figure}[tbh]
	\centering
	\includegraphics[width=0.8\linewidth]{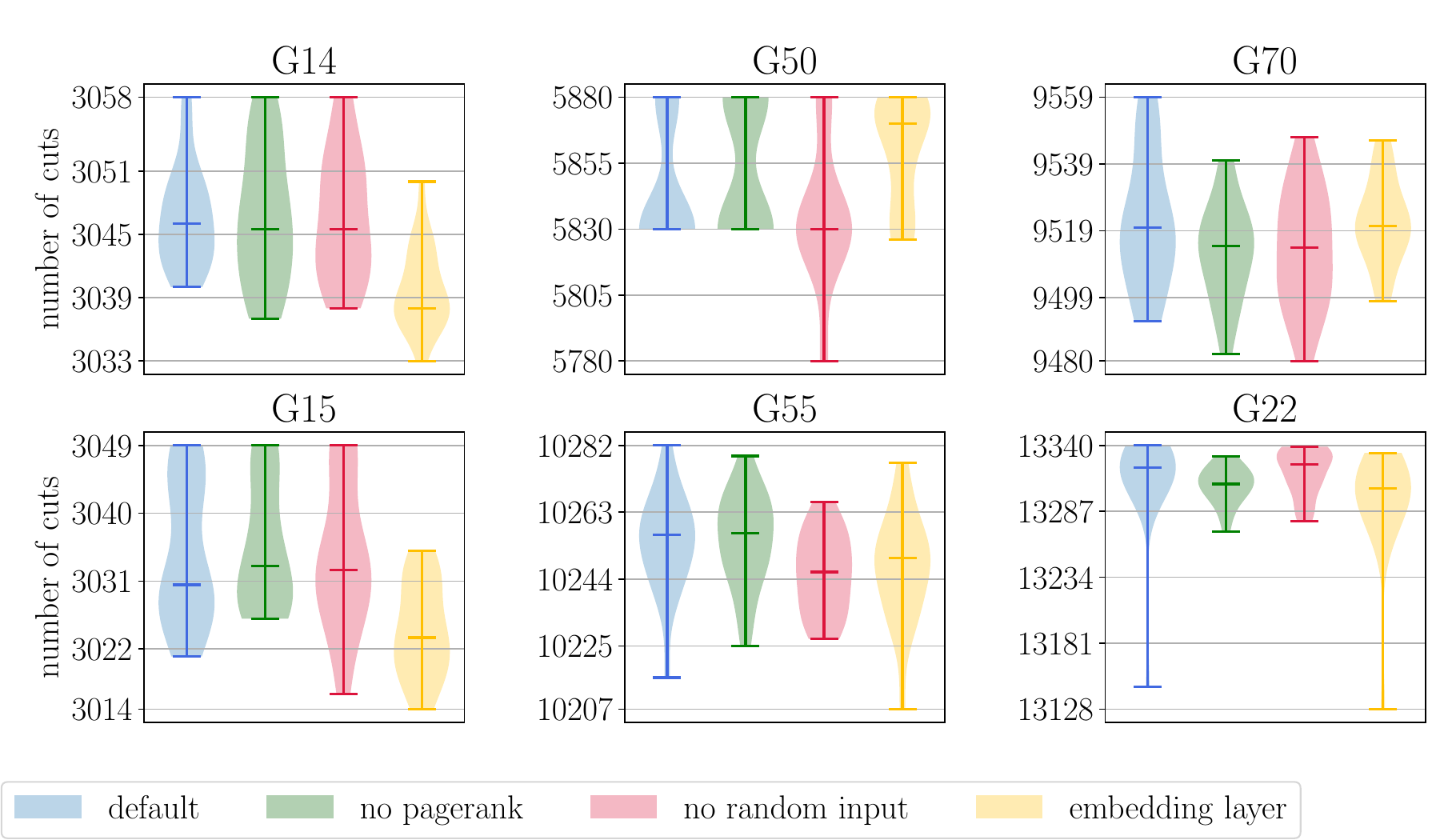}
	\caption{Results distribution for 20 runs of QRF-GNN with the default architecture on several instances from Gset.
	Input feature vectors varied as follows: the default choice corresponds to the blue color;
	the exclusion of the pagerank component corresponds to the green color; the exclusion of the random part corresponds to the red color.
	The use of a trainable embedding layer instead of artificial features is indicated in yellow.
	}
	\label{fig:features}
\end{figure}

In this work, we settled on combining a random vector, shared vector \cite{PosStructFeatures} and pagerank \cite{BRIN1998107} for the input feature vector by default.
Fig. \ref{fig:features} shows the results when one of the parts (random vectors or the pagerank of nodes) was removed.
In some cases, the median improves after dropping features, but the upper bound tends to only go down as the number of input features decreases.
Using an embedding layer does not show any benefit over the default version.

%\begin{table}[h!]
%	\caption{The first row contains the average \emph{P-value} over 200 random regular graphs with $d=5$ for 1 run of QRF-GNN with same configurations as in the figure \ref{fig:layers} and \ref{fig:d5-layers}.
%		 The second row shows the average \emph{P-value} over 200 graphs when the best cut out of 5 runs is taken.}
%\label{table:ablation-maxcut-gset_1conv}
%\centering
%\begin{tabular}{l| l l l l}
% \multicolumn{1}{c|}{\bf Graph} & \multicolumn{1}{c}{\bf Default} & \multicolumn{1}{c}{\bf no SAGEConv (mean)} & \multicolumn{1}{c}{\bf no SAGEConv (pool)} & \multicolumn{1}{c}{\bf SAGEConv (pool), 2\times hid\_dim} \\
% \hline
% & & & &\\
% 	G14 & 3058 &  & 0.725 & \\
% 	G15 & 3049 & 0.732 & 0.725 & \\
%	G22 & 13339 & 0.732 & 0.725 & \\
%	G50 & 5880 & 0.732 & 0.725 & \\
%	G55 & 10282 & 0.732 & 0.725 & \\
%	G70 & 9559 & 0.732 & 0.725 & \\
%\end{tabular}
%\end{table}

\begin{table}[tbh]
	\caption{The first row contains the average \emph{P-value} over 200 random regular graphs with $d=5$ for 1 run of QRF-GNN with same configurations as in Figures \ref{fig:layers} and \ref{fig:d5-layers}.
		 The second row shows the average \emph{P-value} over 200 graphs when the best cut out of 5 runs is taken.}
\label{table:ablation-maxcut-dreg}
\centering
\begin{tabular}{l| l l l}
 \multicolumn{1}{c|}{\bf Runs} & \multicolumn{1}{c}{\bf Default} & \multicolumn{1}{c}{\bf no SAGEConv (mean)} & \multicolumn{1}{c}{\bf no SAGEConv (pool)} \\
 \hline
 & & & \\
 1 & 0.734 & 0.732 & 0.725  \\
 5 & 0.738 & 0.737 & 0.734 \\
\end{tabular}
\end{table}

\end{document}